\documentclass[a4paper,fleqn]{cas-sc}

\usepackage[authoryear,longnamesfirst]{natbib}

\usepackage{comment}
\usepackage{amsmath,amsfonts}
\usepackage{algorithmic}
\usepackage{array}
\usepackage[labelformat=parens,labelsep=space]{subcaption}
\usepackage{textcomp}
\usepackage{stfloats}
\usepackage{url}

\usepackage{tabularray}
\usepackage[section]{placeins}
\usepackage{verbatim}
\usepackage{graphicx}
\usepackage{multirow}
\usepackage[inkscapelatex=false]{svg}
\usepackage{hyphenat}  
\usepackage{diagbox}
\usepackage{footnote}

\usepackage[nolist,nohyperlinks]{acronym}

\usepackage{cleveref}

\usepackage{xurl}
\usepackage{hyperref}
\usepackage{silence}
\begin{document}

\begin{titlepage}
\begin{center}
\vspace*{1cm}

\textbf{\large From Correlation to Causation in Lane Change Prediction for Automated Driving: A Causal Explanation Framework}

\vspace{1.5cm}

Mohamed Manzour$^{1}$ (ahmed.manzour@uah.es), Aditya Kumar$^{1}$ (aditya.kumar@uah.es), Augusto Luis Ballardini$^{1}$ (augusto.ballardini@uah.es) and Miguel \'Angel Sotelo$^{1}$ (miguel.sotelo@uah.es) \\

\hspace{10pt}

\begin{flushleft}
\small  
$^{1}$ Computer Engineering Department, University of Alcalá, Madrid, Spain \\

\vspace{1cm}
\textbf{Corresponding Author:} \\
Mohamed Manzour \\
San Diego Square, s/n. 28801, Alcalá de Henares, Madrid, Spain \\
Tel: +34 685687603 \\
Email: ahmed.manzour@uah.es

\end{flushleft}        
\end{center}
\end{titlepage}

\let\WriteBookmarks\relax
\def\floatpagepagefraction{1}
\def\textpagefraction{.001}
\shorttitle{From Correlation to Causation in Lane Change Prediction for Automated Driving: A Causal Explanation Framework}
\shortauthors{Mohamed Manzour et~al.}

\title[mode = title]{From Correlation to Causation in Lane Change Prediction for Automated Driving: A Causal Explanation Framework}                      

\tnotetext[1] {This work has been funded by the SEGVAUTO5G-CM project of the Regional Government of Madrid under reference 2024/ECO-277 and the PREDATOR project of the Regional Government of Madrid under reference IND2024/IND-35292.}

\author{Mohamed Manzour}[orcid=0009-0007-2009-3573]
\cormark[1]
\ead{ahmed.manzour@uah.es}

\author{Aditya Kumar}[orcid=0009-0004-5717-1027]
\ead{aditya.kumar@uah.es}

\author{Augusto Luis Ballardini}[orcid=0000-0001-6688-5081]
\ead{augusto.ballardini@uah.es}

\author{Miguel {\'A}ngel Sotelo}[orcid=0000-0001-8809-2103]
\ead{miguel.sotelo@uah.es}

\cortext[cor1]{Corresponding author}

\affiliation{organization={Computer Engineering Department, University of Alcalá},
                state={Madrid},
                country={Spain}}

\begin{acronym}[TDMA] 

  \acro{DECI}{Deep End-to-end Causal Inference}
  \acro{TTC}{Time-To-Collision}
  \acro{SVM}{Support Vector Machine}
  \acro{NGSIM}{Next Generation Simulation}
  \acro{ANN}{Artificial Neural Network}
  \acro{XGBoost}{eXtreme Gradient Boosting}
  \acro{HMM}{Hidden Markov Model}
  \acro{LSTM}{Long Short-Term Memory} 
  \acro{RNN}{Recurrent Neural Network} 
  \acro{PREVENTION}{PREdiction of VEhicles iNTentIONs}
  \acro{CNN}{Convolutional Neural Network}
  \acro{GNN}{Graph Neural Network}
  \acro{DyGNN}{Dynamic Graph Neural Network}
  \acro{KGE}{Knowledge Graph Embedding}
  \acro{RAG}{Retrieval Augmented Generation}
  \acro{LLM}{Large Language Model}
  \acro{SCM}{Structural Causal Model}
  \acro{DAG}{Directed Acyclic Graph}
  \acro{ATE}{Average Treatment Effect}

\end{acronym}

\begin{abstract}
Lane-change prediction is a central task in intelligent vehicles, where early and explainable maneuver anticipation can support safer decision-making. However, many existing approaches mainly learn statistical associations between observed driving variables and future maneuvers, while overlooking the causal dependencies among the input variables themselves. This limits interpretability, especially when physically related variables such as longitudinal gap, relative longitudinal velocity, and Time-To-Collision (TTC) are treated as independent flat inputs. This article presents a causal-inference-based framework for lane-change prediction and explanation. The proposed approach combines linguistic feature construction, expert-constrained causal discovery using a constraint matrix, deep structural causal modeling with Deep End-to-end Causal Inference (DECI), intervention-based effect analysis, refutation testing, and recursive causal-chain explanation. The objective is not only to predict the future maneuver, but also to identify candidate variables that directly contribute to the prediction, the upstream factors influencing them, and the causal chains through which these effects propagate. The framework is evaluated for three maneuver classes: left lane change (LLC), lane keeping (LK), and right lane change (RLC). The model achieves average F1-scores above 95\% during the first three seconds before the lane-marking crossing event, while still providing maneuver anticipation in the 7--8 s interval before the lane-change event. Beyond prediction accuracy, the framework uses intervention-based effect analysis to distinguish influential from weakly influential variables under the learned causal structure. It further distinguishes candidate direct contributors from mediated effects and generates contrastive causal-chain explanations that clarify why the predicted maneuver is favored and why the alternative maneuvers are less supported. The main contribution is therefore a mechanism-aware lane-change prediction pipeline that moves beyond correlation-based classification toward more interpretable causal reasoning for maneuver prediction.
\end{abstract}


\maketitle

\section{Introduction}
\label{sec:introduction}
\begin{figure}[pos=!htbp]
\centering
\includegraphics[width=\columnwidth]{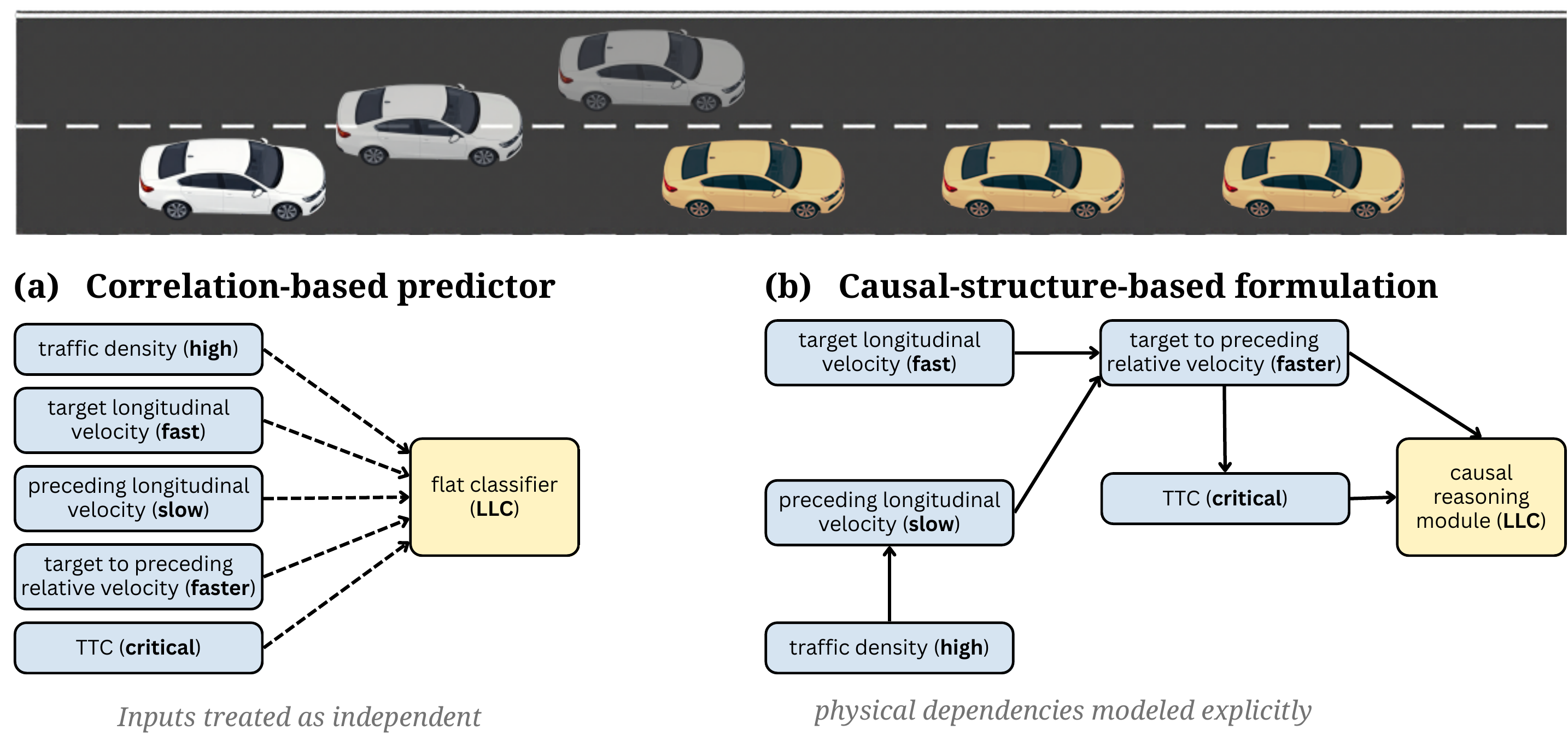}
\caption{Scenario-based motivation for causal lane-change prediction. The white target vehicle encounters dense traffic and a slower preceding vehicle in its lane. This creates a high relative longitudinal velocity and a critical \ac{TTC}, leading to an anticipated left lane change, shown by the semi-transparent future positions of the target vehicle. In the correlation-based formulation, all variables are treated as independent inputs directly connected to the maneuver classifier, which may hide whether the prediction is mainly driven by TTC, relative velocity, vehicle speeds, or traffic density. In the causal-structure-based formulation, the variables are organized according to their physical dependency path, enabling the prediction to be explained as a causal chain from upstream traffic conditions to the final maneuver decision.}
\label{fig:intro}
\end{figure}

Lane-change prediction is a fundamental task in intelligent vehicles and advanced driver-assistance systems. Anticipating whether a surrounding vehicle will keep its lane, change to the left lane, or change to the right lane allows an automated vehicle to prepare safer and smoother decisions in advance. This is particularly important in dense and dynamic traffic, where delayed or incorrect maneuver anticipation may lead to unsafe braking, uncomfortable control actions, or insufficient time for cooperative responses. Recent lane-change prediction approaches have achieved promising performance using machine learning and deep learning models trained on vehicle trajectory and interaction features. These models commonly use variables such as vehicle speed, relative velocity, longitudinal gap, time headway, \ac{TTC}, and surrounding traffic density to predict future maneuvers. However, most of these approaches are primarily correlation-driven: they learn statistical associations between input variables and maneuver classes, but they do not explicitly model the physical and causal dependencies among the variables themselves. As a result, the model may predict the correct maneuver while still providing an incomplete or ambiguous explanation of why this maneuver was predicted. This limitation becomes critical because many commonly used lane-change features are not independent. For example, \ac{TTC} with the preceding vehicle is not an isolated driving variable; it is affected by upstream interaction factors such as the relative longitudinal velocity between the target vehicle (the vehicle whose future maneuver is being predicted) and preceding vehicle (the vehicle in front of the target vehicle), which itself depends on the velocities of both vehicles. Therefore, if a prediction model strongly relies on \ac{TTC}, this does not necessarily mean that \ac{TTC} is always the original reason behind the maneuver. In some cases, it may be the main direct contributor; in other cases, it may only mediate the effect of upstream variables such as vehicle speed differences or traffic conditions. A correlation-based model may hide this distinction because it mainly captures which variables are statistically associated with the output, rather than how their effects propagate through the driving interaction. \Cref{fig:intro} illustrates the motivation using a representative lane-change scenario. The target vehicle, highlighted in white, travels in a lane where the surrounding traffic is relatively dense, while the neighboring vehicles are shown in a different color. As the target vehicle proceeds, it approaches a slower preceding vehicle in the same lane. This interaction produces a higher relative longitudinal velocity between the target and preceding vehicles, which in turn leads to a critical \ac{TTC}. The semi-transparent positions of the target vehicle indicate the anticipated future maneuver, where the vehicle is expected to perform a left lane change as a consequence of this interaction sequence. Therefore, the prediction is not interpreted as the result of one isolated variable, such as \ac{TTC} alone, but as the outcome of a causal chain that starts from the traffic context and propagates through vehicle-speed interaction variables toward the final maneuver prediction.
This example also highlights the limitation of conventional correlation-based predictors. As shown in \Cref{fig:intro}(a), a flat classifier receives traffic density, target velocity, preceding-vehicle velocity, relative velocity, and \ac{TTC} as independent inputs directly connected to the predicted maneuver. Such a model may learn that critical \ac{TTC} is strongly correlated with a left lane change, but it cannot explain whether \ac{TTC} is the main reason in that case, whether it mediates the effect of upstream variables, or whether the maneuver is mainly driven by another factor such as the speed difference or the traffic density.
In contrast, \Cref{fig:intro}(b) represents the variables according to their assumed physical dependencies, allowing the prediction to be interpreted through a chain of upstream and intermediate factors rather than through isolated feature associations. This conceptual difference highlights that predictive correlation does not necessarily imply direct causal influence.

Motivated by this gap, this article investigates lane-change prediction from a causal-inference perspective. Instead of using the model only to classify future maneuvers, the proposed framework aims to identify which variables directly contribute to the prediction, which upstream factors influence them, and how these effects propagate through causal chains toward the predicted maneuver. In addition, the framework uses intervention-based effect analysis to examine whether changing a given feature produces a meaningful change in the predicted maneuver under the learned causal structure. This allows the model to distinguish variables that are influential for the maneuver prediction from variables whose effect is weak or negligible. The objective is not to claim definitive real-world causality, but to provide a mechanism-aware prediction framework that can distinguish direct contributors, mediating variables, upstream causes, and weakly influential factors under the learned causal structure.

The proposed framework combines linguistic feature construction, expert-constrained causal discovery, deep structural causal modeling using \ac{DECI}, intervention-based effect analysis, refutation testing, and recursive causal-chain explanation. Expert knowledge is incorporated through a constraint matrix that restricts implausible causal directions and prevents the model from learning relationships that contradict the physical structure of the driving scenario. Intervention analysis is then used to estimate the influence of candidate variables on the predicted maneuver, while refutation testing evaluates the stability of these estimated effects under perturbations. The framework is evaluated for three maneuver classes: Left Lane Change (LLC), Lane Keeping (LK), and Right Lane Change (RLC). Beyond prediction accuracy, the proposed approach explains why the predicted maneuver is favored while alternative maneuvers are less supported.

To support the reading of the following sections, the terminology used in this article is first made
explicit. The target vehicle (shown in white in all figures) is the vehicle whose future maneuver is
being predicted; every driving variable and explanation is defined from its perspective. The
surrounding vehicles are named according to their position relative to the target vehicle. The
preceding vehicle is the vehicle directly ahead of the target in the same lane, and the following
vehicle is the one directly behind it in the same lane. Vehicles in the adjacent lanes are named
analogously: the left-preceding and right-preceding vehicles are ahead of the target in the left and
right adjacent lanes, while the left-following and right-following vehicles are behind it in those
lanes. A vehicle traveling next to the target in an adjacent lane is referred to as an alongside (left-alongside or right-alongside) vehicle. Interaction variables such as longitudinal gap, relative longitudinal velocity, and TTC are always
computed between the target vehicle and a specified neighboring vehicle; for example, the precedingVehicle TTC denotes the TTC between the target and its preceding vehicle. This convention is used
consistently in all subsequent sections, figures, and causal-chain explanations.
\section{Literature and Research Foundation}
\label{sec:related-work}

\subsection{Literature Review}

Lane-change prediction has been widely studied in intelligent vehicles and autonomous driving. Existing approaches can be broadly grouped into rule-based models, traditional machine learning models, deep learning models, graph-based models, and more recent explainable or knowledge-driven approaches.

Early lane-change prediction methods relied on rule-based formulations, such as gap-acceptance models, where a lane change is assumed to occur when the target vehicle satisfies predefined conditions related to the available gap with front and rear vehicles in the current or adjacent lanes \cite{yang1996microscopic}. Although these models are interpretable, their performance is usually limited by the manually defined rules and their ability to represent complex interaction patterns. Data-driven methods were therefore introduced to learn the mapping between observed driving variables and future maneuvers from recorded traffic data.

Several works have used traditional machine learning models for lane-change prediction. For example, \cite{woo2017lane} combined \acp{SVM} and artificial potential field models to detect lane changes using the \ac{NGSIM} dataset, considering features such as distance from the lane centerline, lateral velocity, and potential-field-based information. Similarly, \cite{benterki2019prediction} trained \acp{SVM} and \acp{ANN} models using kinematic variables, including longitudinal and lateral velocities, accelerations, distances to lane markings, yaw angle, and yaw rate. Other works adopted models such as \ac{XGBoost}, logistic regression, random forests, \acp{HMM}, and Bayesian-network-based approaches to predict lane-change intention from numerical trajectory and interaction features \cite{izquierdo2017vehicle,yuan2018lane,xia2021human,li2023early,zhang2022xgboost,syama2022ensemble,ma2021lane,li2019dynamic,liu2024peril}. These methods showed that trajectory and surrounding-vehicle variables are useful for predicting lane-change maneuvers, but they mostly treat the inputs as flat predictors of the output class.

Deep learning methods have also been extensively investigated due to their ability to model temporal driving patterns. \cite{su2018learning} employed a \ac{LSTM} model using the target vehicle trajectory and surrounding-vehicle states extracted from the \ac{NGSIM} dataset. Their inputs included the vehicle's lateral and longitudinal positions, acceleration, surrounding-vehicle existence indicators, and longitudinal distances to neighboring vehicles. \cite{laimona2020implementation} and \cite{amer2024enhancing} both investigated recurrent models for vehicle-intention prediction using the \ac{PREVENTION} dataset. While \cite{laimona2020implementation} compared \acp{RNN} and \acp{LSTM} using sequences of bounding-box centroid coordinates, \cite{amer2024enhancing} developed a three-layer \ac{LSTM} model based on using center-point, distance, and yaw-angle features. Image-based and hybrid models were also explored; for instance, \cite{izquierdo2019experimental} proposed Motion History Image--\ac{CNN} and GoogLeNet--\ac{LSTM} architectures using visual information from the \ac{PREVENTION} dataset, while other works used two-stream networks, action-recognition architectures, and efficient environment representations for lane-change classification and prediction \cite{fernandez2020two,izquierdo2021vehicle,liang2022lane,biparva2022video}. These works demonstrate the value of temporal and visual representations, but the learned reasoning process remains difficult to interpret.

More recent approaches have integrated lane-change intention prediction with trajectory prediction. \cite{xue2022integrated} used \ac{XGBoost} for lane-change prediction and \ac{LSTM} for trajectory prediction on the highD dataset, considering traffic density, vehicle type, and relative trajectory features between the target and surrounding vehicles. \cite{gao2023dual} proposed a dual-transformer architecture in which one transformer predicts lane-change intention and the second transformer predicts the future trajectory by combining intention information with historical lateral motion.
Interaction-aware models have also been explored using graph-based and transformer-based architectures. For example, \cite{li2023social,lu2025lane} modeled inter-vehicle interactions using \acp{GNN}, while \cite{guo2025vehicle,lu2025knowledge} employed transformer-based models for lane-change prediction using surrounding-vehicle states. More recently, \cite{wu2026dynamic} proposed a \ac{DyGNN} combined with a Transformer to capture time-varying inter-vehicle interactions for trajectory prediction and driving-intention recognition. These methods show the increasing importance of interaction modeling in lane-change prediction; however, the learned relations are mainly used to improve predictive performance rather than to explicitly estimate causal feature influence or generate causal-chain explanations.

In parallel, several works have focused on improving interpretability and knowledge integration. Knowledge-based and neuro-symbolic approaches have been proposed to represent driving contexts using linguistic inputs and structured relations. For example, \cite{manzour2024vehicle} used \acp{KGE} and Bayesian inference to predict safe lane changes from linguistic features such as lateral velocity, lateral acceleration, and \ac{TTC} risk. This direction was later extended by integrating \ac{RAG} to provide natural-language explanations grounded in external knowledge \cite{hussien2025rag,manzour2025explainable}. \acp{LLM} have also been explored for explainable lane-change intention and trajectory prediction, where textual prompts are used to improve the explainability of the prediction output \cite{peng2025lc}. Recent works have also moved toward knowledge-enhanced and physics-informed prediction. \cite{lu2026knowlcp} proposed KnowLCP, a Knowledge-Augmented Lane-Change Prediction framework that integrates driving knowledge into intention recognition and trajectory forecasting. Similarly, \cite{shi2026multiscenario} proposed a physics-informed framework for three-class lane-change intention prediction, incorporating vehicle kinematics, interaction feasibility, and traffic-safety indicators such as headway, \ac{TTC}, and gap-related measures. Although these approaches improve interpretability, knowledge usage, or physical consistency, they still do not explicitly learn an expert-constrained \ac{SCM}, perform intervention-based feature influence analysis, apply refutation-based robustness checks, or generate contrastive causal-chain explanations.

Overall, the literature shows that lane-change prediction has progressed from rule-based models to machine learning, deep learning, transformer-based, graph-based, knowledge-enhanced, and explainable approaches. These works have demonstrated the importance of trajectory features, surrounding-vehicle interactions, traffic context, linguistic representations, and physically informed indicators such as time headway and \ac{TTC} for maneuver prediction.

Causal inference has also been widely studied as a framework for moving beyond statistical association toward estimating the effect of interventions under explicit structural assumptions \cite{pearl2009causal}. It has been used in domains such as epidemiology, medicine, economics, and policy evaluation, where the objective is not only to predict an outcome, but also to understand whether changing a given variable would produce a meaningful effect on that outcome \cite{athey2017state}. Recent machine-learning-based causal frameworks, such as \ac{DECI}, further combine causal discovery and causal inference from observational data, enabling causal graph learning and treatment-effect estimation within a unified framework \cite{geffner2022deep}.

\subsection{Research Gap}

Despite the progress achieved in lane-change prediction, most existing approaches still formulate the problem mainly as a correlation-based classification task. Traditional machine learning, deep learning, transformer-based, graph-based, knowledge-enhanced, and recent \ac{LLM}-based models commonly learn associations between trajectory, interaction, and traffic variables and the future maneuver class. However, they do not explicitly model the causal dependencies among the input variables themselves. This is limiting because several commonly used predictors, such as longitudinal gap, relative longitudinal velocity, time headway, and \ac{TTC}, are physically related rather than independent. As a result, existing models may achieve high prediction accuracy while still being unable to distinguish whether a variable is a direct contributor, an upstream cause, a mediator, or only a weakly influential feature under a learned causal structure.

A second gap concerns the limited use of intervention-based and robustness-based analysis in lane-change prediction. Existing explainable approaches may provide interpretable inputs, feature importance scores, or natural-language explanations, but they generally do not test whether changing a given feature produces a meaningful effect on the predicted maneuver under a causal structure. Similarly, refutation tests are rarely used to assess whether the estimated effects remain stable under perturbations, such as placebo treatments, random common causes, or data-subset changes.

A third gap is the limited integration of expert driving knowledge into the causal discovery process. In safety-critical driving scenarios, some causal directions are physically implausible and should be prevented during structure learning. However, most existing approaches rely mainly on observational data without explicitly restricting the learning process using domain knowledge. This motivates an expert-constrained causal framework that prevents implausible relationships from being learned while allowing the remaining dependencies to be discovered from data.

\subsection{Research Objective}

This work aims to develop a mechanism-aware lane-change prediction framework that moves beyond correlation-based classification toward causal reasoning, intervention-based effect analysis, and causal-chain explanation. The objective is not to claim definitive real-world causality, but to learn and analyze an expert-constrained structural causal representation that supports prediction, feature influence analysis, robustness checking, and explanation. The specific objectives are:

\begin{enumerate}
    \item \textbf{Integrate expert knowledge into causal structure learning:}
    The proposed framework incorporates expert driving knowledge through a constraint matrix that specifies causal directions that should not be learned because they contradict the physical logic of the driving scenario. The \ac{DECI} model then learns the remaining structural dependencies from observational driving data under these constraints. In this way, the learned causal structure is not purely data-driven or manually fixed, but expert-constrained and data-adaptive.

    \item \textbf{Estimate feature influence through intervention analysis:}
    Intervention-based effect analysis is used to examine whether changing a given feature produces a meaningful influence on the predicted maneuver under the learned causal structure. This allows the framework to distinguish influential variables from weakly influential variables.

    \item \textbf{Evaluate the stability of the estimated effects:}
    Refutation tests are applied to assess whether the estimated effects remain stable under perturbations, such as placebo treatments, random common causes, or data-subset changes.

    \item \textbf{Generate contrastive causal-chain explanations:}
    The learned structure is used to trace causal chains from upstream variables through intermediate factors toward the predicted maneuver, explaining why the predicted maneuver is favored and why alternative maneuvers are less supported.
\end{enumerate}

\section{Methodology}
\label{sec:methodology}

\subsection{Framework Overview}
\label{subsec:framework-overview}

\begin{figure}[pos=!htbp]
\centering
\includegraphics[width=\columnwidth]{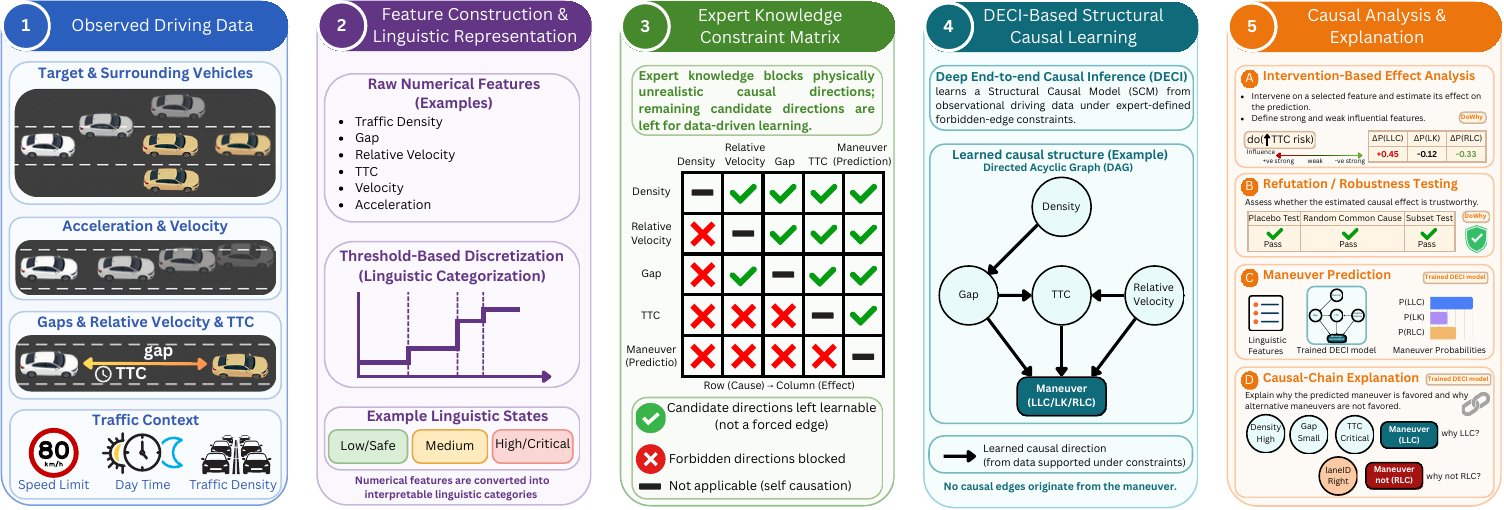}
\caption{Overall methodology of the proposed causal-inference-based lane-change prediction and explanation framework. The pipeline starts from observed driving data, including target-vehicle motion, surrounding-vehicle interactions, gap-related variables, TTC, and traffic context. The numerical variables are converted into interpretable linguistic categories using thresholds derived from the literature and from dataset observation. Expert knowledge is then incorporated through a constraint matrix that blocks physically unrealistic causal directions while leaving the remaining candidate directions available for data-driven learning. Under these constraints, DECI learns an \ac{SCM} represented as a DAG. The learned model is then used for causal analysis and explanation: intervention-based effect analysis and refutation testing are performed using DoWhy, while the trained DECI-based \ac{SCM} is used to estimate maneuver probabilities and support contrastive causal-chain explanations.}
\label{fig:methodology}
\end{figure}

The proposed methodology is designed to predict lane-change maneuvers while also explaining the causal mechanisms that support the prediction. The proposed framework first represents the driving variables in an interpretable causal structure. Based on this structure, the framework supports maneuver prediction, causal-effect analysis, and robustness evaluation. Then, it examines not only which maneuver is predicted, but also which variables influence the prediction, and how these variables are connected together.
The overall pipeline is illustrated in \Cref{fig:methodology}. The framework starts from observed driving data describing the target vehicle, surrounding vehicles, traffic context, and interaction-related variables. These variables include kinematic information, gap-related measures, relative velocity, \ac{TTC}, and other safety-relevant indicators. Then, the numerical features are converted into interpretable linguistic categories using thresholds derived from the literature and from empirical observation of the dataset distributions. This representation allows the model to reason over human-understandable driving states.
After constructing the linguistic feature representation, expert driving knowledge is incorporated into the causal structure learning process through a constraint matrix. This matrix specifies causal directions that should not be learned because they contradict the temporal, physical, or definitional logic of the driving scenario. The constraint matrix does not force the model to learn specific causal links; rather, it blocks physically unrealistic directions and leaves the remaining candidate directions available for data-driven learning. In this way, the causal discovery process is neither fully manual nor purely data-driven, but expert-constrained and data-adaptive. \ac{DECI} is then used to learn an \ac{SCM} from the observational driving data under the defined expert constraints. The learned structure is represented as a \ac{DAG}. A \ac{DAG} is a directed graph in which the edges have a direction, but no directed cycle is allowed. This means that the graph can contain causal paths such as $X_1 \rightarrow X_2 \rightarrow Y$, but it cannot contain a feedback loop such as $X_1 \rightarrow X_2 \rightarrow X_1$, where following the arrows returns to the starting variable. In this graph, nodes correspond to the linguistic driving variables and the maneuver outcome, while directed edges represent the learned causal dependencies among them. Since the maneuver corresponds to a future prediction outcome, no causal edges are allowed to originate from the maneuver node toward the observed input features.

The final stage uses the learned \ac{SCM} for causal analysis and explanation. Intervention-based effect analysis is used to examine whether changing a selected feature produces a meaningful effect on the predicted maneuver. Refutation tests are then applied to assess the robustness of the estimated effects using placebo treatment, random common cause, and subset tests. The trained DECI-based model is also used to estimate the probability of each maneuver class, and the class with the highest probability is selected as the predicted maneuver. Finally, the learned causal structure is recursively traced to generate contrastive causal-chain explanations that clarify why the predicted maneuver is favored and why alternative maneuvers are less supported.

\subsection{Feature Construction and Linguistic Categorization}
\label{subsec:feature-construction}

The highD naturalistic highway trajectory dataset is used as the source of the observed driving variables. The dataset contains drone-recorded vehicle trajectories on German highways, making it suitable for studying lane-change behavior in dense and realistic traffic conditions \cite{krajewski2018highd}. The original numerical feature file contains trajectories of approximately $110,000$ vehicles, from which interaction, lane-context, and traffic-context variables are extracted. Training samples are extracted $5$ seconds before the lane-change event, where the event is defined as the instant when the target vehicle reaches the lane marking during the maneuver. To avoid data leakage, the training and test sets are separated at the track/recording level rather than by randomly splitting individual samples. In other words, samples from the same recording are not shared between training and testing. The experimental setup uses a $75/25$ split, where the training portion is used to fit the DECI model and the held-out portion is used only for test evaluation. After class balancing, the training set contains $450,000$ samples/frames, with $150,000$ samples for each maneuver class. The test set contains $150,000$ samples, with $50,000$ samples for each class. Therefore, the experimental data used for training and testing contain $600,000$ samples in total.

The numerical data are then converted into linguistic features before being used by the DECI-based causal model. This representation serves two purposes. First, it converts continuous driving measurements into finite categorical states suitable for the categorical formulation used by the causal model. Second, it aligns the input space with interpretable driving concepts, allowing the learned causal relationships to be described using human-understandable states such as \textit{safe}, \textit{critical}, \textit{close}, \textit{far}, \textit{slow}, or \textit{fast}. The selected variables are converted into categorical string-valued features using threshold-based discretization. The final input to the DECI-based causal model consists of 47 linguistic variables and the maneuver label. The selected features cover target vehicle kinematics, including both longitudinal and lateral motion, lane and road context, surrounding-vehicle occupancy, neighboring-vehicle interaction measures, and traffic-density information. \Cref{tab:linguistic_features} summarizes the feature groups and their linguistic representations.
\begin{table*}[pos=!htbp]
\centering
\caption{Summary of the linguistic feature groups used as input to the DECI-based causal model.}
\label{tab:linguistic_features}
\begin{tabular}{|p{2.5cm}|p{5.5cm}|p{6.9cm}|}
\hline
\textbf{Feature group} & \textbf{Representative variables} & \textbf{Linguistic representation} \\
\hline

Target vehicle kinematics
& Longitudinal speed, longitudinal acceleration, lateral velocity, lateral acceleration
& Speed, acceleration, and lateral-motion variables are converted into ordered motion states such as \textit{very\_slow}, \textit{fast}, \textit{hard\_braking}, \textit{moving\_left}, and \textit{moving\_right}. \\
\hline

Lane and road context
& Lane rank, posted speed limit, rush-hour flag
& Road-context variables are encoded using categorical states such as \textit{rightmost\_lane}, \textit{center\_lane}, \textit{leftmost\_lane}, \textit{no\_limit}, \textit{130\_kmh}, and binary rush-hour states. \\
\hline

Surrounding-vehicle occupancy
& Presence of left/center/right preceding/following, left-side, and right-side neighboring vehicles
& Neighbor availability is encoded using binary states. If a neighboring vehicle is absent, its related interaction features are encoded as \textit{not\_applicable}. \\
\hline

Neighbor interaction features
& Longitudinal gap, neighbor speed, relative longitudinal velocity, and TTC for surrounding vehicles
& Interaction variables are represented using ordered states such as \textit{very\_close}, \textit{close}, \textit{far}, \textit{much\_slower}, \textit{similar}, \textit{much\_faster}, \textit{critical}, and \textit{safe}. \\
\hline

Traffic context
& Ego-lane, left-lane, and right-lane traffic density; ego speed-to-limit ratio
& Traffic and speed-context variables are encoded using states such as \textit{empty}, \textit{low}, \textit{medium}, \textit{congested}, \textit{well\_below}, \textit{near\_limit}, and \textit{well\_above}. \\
\hline

Maneuver label
& Future maneuver class
& The prediction target is encoded as one of three classes: \textit{LLC}, \textit{LK}, or \textit{RLC}. \\
\hline

\end{tabular}
\end{table*}
The numerical-to-linguistic conversion is performed using threshold-based discretization, with category boundaries derived from common practice in the traffic-safety literature and from empirical observation of the highD dataset distributions \cite{krajewski2018highd}, following a categorization approach similar to our previous work \cite{manzour2024vehicle, manzour2025explainable}. As a representative example, TTC is discretized using boundaries at $-10$, $0$, $4$, and $15$ s, where negative values indicate diverging interactions, the $4$ s boundary represents a commonly adopted critical interaction level \cite{saffarzadeh2013general, ramezani2020comparing, manzour2024vehicle}, and values above $15$ s are treated as non-interacting based on the empirical TTC distribution of the dataset. The remaining variables are discretized in the same manner. Since the linguistic categorization is a preprocessing step and not the main focus of this work, the proposed causal framework can operate over any categorical feature representation, and the investigation of alternative discretization and categorization methods is left for future work.

\subsection{Expert Knowledge Constraint Matrix}
\label{subsec:constraint-matrix}

After constructing the linguistic feature representation, expert driving knowledge is incorporated into the causal structure learning process through a constraint matrix. This matrix defines the search space of the causal discovery model by specifying which directed relations are not allowed to be learned. The objective is not to manually impose a complete causal graph, but to prevent the model from selecting directions that contradict the temporal, physical, or definitional logic of the driving scenario. As illustrated in \Cref{fig:methodology}, an allowed entry does not represent a forced causal edge; it only indicates that the corresponding direction is left available for data-driven learning if supported by the observational data.

The construction of the constraint matrix was supported by two sources of prior knowledge: expert driving-domain reasoning and LLM-assisted causal-direction review using PyWhy-LLM \cite{kiciman2023causal}. PyWhy-LLM is a prompt-based causal-analysis support tool that uses external \acp{LLM} to suggest or critique causal relationships from natural-language variable descriptions. In this work, it was used as an auxiliary review tool over linguistic and descriptive driving concepts, such as traffic density, rush-hour context, speed-limit context, longitudinal gap, relative velocity, TTC, and maneuver outcome. The LLM output was not used to prove causality or to force edges into the graph. Instead, it supported expert inspection of candidate directions, while the final constraint decisions remained expert-defined. For example, PyWhy-LLM was queried using linguistic driving concepts to compare the candidate directions between traffic density and rush-hour context. The prompt asked whether traffic density causes rush hour, rush hour causes traffic density, or neither relation is more plausible. The model selected the second option and reasoned that rush hour is a temporal traffic condition that can increase the number of vehicles entering the road network, thereby increasing traffic density, whereas traffic density does not determine the rush-hour state. This output supported the expert decision to block the reverse direction from traffic density to rush-hour context, while leaving the direction from rush-hour context to traffic density available for DECI to learn from data. Importantly, this LLM-assisted review only supported the identification of directions that should be excluded from the causal search space.

The constraint rules follow several types of domain knowledge. First, temporal constraints are imposed on the maneuver outcome. Since the maneuver represents a future prediction target, it is not allowed to cause any observed input feature. For example, the future maneuver class cannot cause current traffic density, longitudinal gap, relative velocity, TTC, lateral velocity, or surrounding-vehicle states. In contrast, observed driving variables are allowed to influence the maneuver outcome if the learned structure supports such relations. Second, contextual variables are constrained according to their role in the driving scene. For example, rush-hour context may influence traffic density because the time of day can affect congestion patterns, but traffic density cannot cause the rush-hour state. Similarly, road-context variables such as lane rank and speed-limit information are treated as upstream contextual conditions. Therefore, instantaneous interaction variables, such as TTC, relative velocity, or traffic density, are not allowed to cause the posted speed limit or the lane-rank context. However, contextual variables can remain available as candidate causes of downstream traffic or interaction variables when such directions are physically meaningful. Third, definitional and physical dependencies are used to restrict directions among interaction variables. A representative case is TTC. TTC is computed from the longitudinal gap and the relative  velocity under closing or diverging conditions. Therefore, directions from longitudinal gap and relative velocity toward TTC are physically meaningful candidate directions, while reverse directions, such as TTC causing the gap or TTC causing the relative velocity, are blocked. This prevents the model from learning causal directions that invert the physical construction of safety-related variables. A similar logic applies to relative velocity, which is derived from the velocities of the interacting vehicles; the derived relative quantity should not be treated as the original cause of the observed speed components. When the relation between two variables is physically possible but uncertain, the corresponding direction is not blocked and left available for DECI to learn from the observational data. Thus, the constraint matrix removes only directions judged to be temporally, physically, or definitionally inconsistent, while preserving a data-adaptive causal discovery process for the remaining candidate dependencies. The resulting learned structure is therefore expert-constrained, but not manually fixed.

\subsection{DECI-Based Structural Causal Learning}
\label{subsec:deci-learning}

After preparing the linguistic feature set and defining the expert constraint matrix, Deep End-to-end Causal Inference (DECI) is used to learn the \ac{SCM} from the observational driving data. DECI is a deep \ac{SCM} that jointly learns a directed causal graph and the associated structural mechanisms from observational data under explicit modeling assumptions \cite{geffner2022deep}. In the proposed framework, DECI receives a categorical data table, where each row represents one traffic sample and each column represents either one linguistic driving variable or the maneuver label.

Let $\mathbf{X}=\{X_1,\ldots,X_d\}$ denote the set of linguistic driving variables, with $d=47$, and let $Y$ denote the maneuver outcome. The complete set of variables used for causal structure learning is defined as:
\begin{equation}
\mathbf{Z} = \mathbf{X} \cup \{Y\}.
\end{equation}
Therefore, each variable $Z_j \in \mathbf{Z}$ represents either one linguistic driving feature or the maneuver label. DECI describes each variable using the following structural form:
\begin{equation}
Z_j = f_j\left(\mathrm{Pa}_{\mathcal{G}}(Z_j), \epsilon_j\right),
\end{equation}
where $Z_j$ is the variable being modeled, and $\mathrm{Pa}_{\mathcal{G}}(Z_j)$ denotes its parents in the learned graph $\mathcal{G}$. The parents of a variable are the variables that have directed edges pointing toward it. The function $f_j(\cdot)$ represents the learned relationship between a variable and its parents. In DECI, this function is learned using neural networks, which allows the model to capture nonlinear relationships without assuming a fixed equation in advance. The term $\epsilon_j$ is included because the learned parents of a variable usually cannot explain that variable perfectly. In real traffic data, the observed variables may miss some factors, such as measurement uncertainty, or values located close to the discretization thresholds. In a numerical setting, this idea is straightforward. For example, if TTC were modeled as a continuous value using longitudinal gap and relative velocity, a deterministic model without noise would write:
\begin{equation}
\mathrm{TTC} = f(\mathrm{gap}, \mathrm{relative\ velocity}).
\end{equation}
This means that the same gap and relative velocity would always produce exactly the same TTC value. However, this is usually unrealistic because two traffic situations with similar observed values may still lead to slightly different TTC values. Therefore, the noise term is added:
\begin{equation}
\mathrm{TTC} = f(\mathrm{gap}, \mathrm{relative\ velocity}) + \epsilon.
\end{equation}
In this numerical case, $\epsilon$ can be understood as the remaining numerical difference between the value predicted from the observed parents and the actual observed value. In this work, however, the variables are linguistic categories rather than continuous numerical values. Therefore, the noise term cannot be added directly to the category. For example, the model cannot compute ``\textit{critical} + noise''. Without uncertainty, the categorical mechanism would behave deterministically. For instance, if the parent states are \textit{gap = small} and \textit{relative velocity = target\_is\_faster}, the model would always assign:
\begin{equation}
P(\mathrm{TTC}=\textit{critical})=1.00,\quad
P(\mathrm{TTC}=\textit{cautious})=0.00,\quad
P(\mathrm{TTC}=\textit{safe})=0.00.
\end{equation}
This would mean that the same parent categories always produce the category \textit{critical}. When uncertainty is considered, the same parent states can instead produce a probability distribution over the possible TTC categories:
\begin{equation}
P(\mathrm{TTC}=\textit{critical})=0.72,\quad
P(\mathrm{TTC}=\textit{cautious})=0.20,\quad
P(\mathrm{TTC}=\textit{safe})=0.08.
\end{equation}
Here, $\epsilon_j$ should be understood as the hidden variation that prevents the observed parent categories from determining one output category with complete certainty. It is not added to the category itself. Rather, its effect appears through the probability distribution: the parent states make \textit{critical} the most likely TTC category, but other categories remain possible because some relevant information is not fully captured by the observed parents. The learning process can be summarized as follows. First, DECI considers possible \ac{DAG} structures among the variables in $\mathbf{Z}$. For notation, let $\mathbf{A}$ denote the edge matrix of the learned graph, where $A_{ij}=1$ means that the directed edge $Z_i \rightarrow Z_j$ is present, and $A_{ij}=0$ means that this edge is absent. The constraint is written as:
\begin{equation}
A_{ij}=0 \quad \text{whenever} \quad C_{ij}=0.
\end{equation}
This means that if the constraint matrix $C$ marks a direction as forbidden, DECI cannot learn that edge. After applying these constraints, DECI learns the graph structure and the functions $f_j$ from the observational driving data. In other words, the model learns both which variables are connected and how each variable depends on its learned parents. After training, the graph forms an \ac{SCM} that can be used for later analysis.

The trained DECI model provides the structural backbone for the remaining stages of the framework. Using the learned graph and structural mechanisms, the framework first performs intervention-based effect analysis to estimate how selected feature-category changes affect the maneuver distribution. DoWhy library is then used to apply refutation tests in order to assess the robustness of the estimated effects. After this causal-effect analysis stage, the trained DECI model is used to estimate maneuver probabilities and to trace causal chains explaining why one maneuver is favored over the alternative maneuvers. Finally, since this article uses DECI as an existing causal learning model rather than proposing a modification of the DECI algorithm, further details about its probabilistic formulation and training procedure are provided in the original work by \cite{geffner2022deep}.

\subsection{Causal Role Interpretation for Intervention and Explanation}
\label{subsec:causal-roles}

After learning the structural causal graph, the next step is to define how this graph will be used in the following causal-analysis stage. The learned graph is not only a visualization of dependencies among variables; it also determines how variables are interpreted during intervention analysis, refutation testing, and causal-chain explanation. This subsection therefore connects the learned graph from \Cref{subsec:deci-learning} with the causal-analysis procedures in \Cref{subsec:causal-analysis} by defining the practical causal roles used in the rest of the methodology. These roles are introduced to answer three questions. First, when estimating an intervention effect, which variable is being changed and which variable is being measured? Second, when explaining a maneuver prediction, which variables are directly connected to the maneuver and which variables act through intermediate paths? Third, which upstream variables may influence both a selected treatment and the outcome and should therefore be considered during effect estimation? Answering these questions is necessary before applying the intervention and explanation procedures in \Cref{subsec:causal-analysis}.

\begin{figure}[pos=!htbp]
\centering
\includegraphics[width=0.90\linewidth]{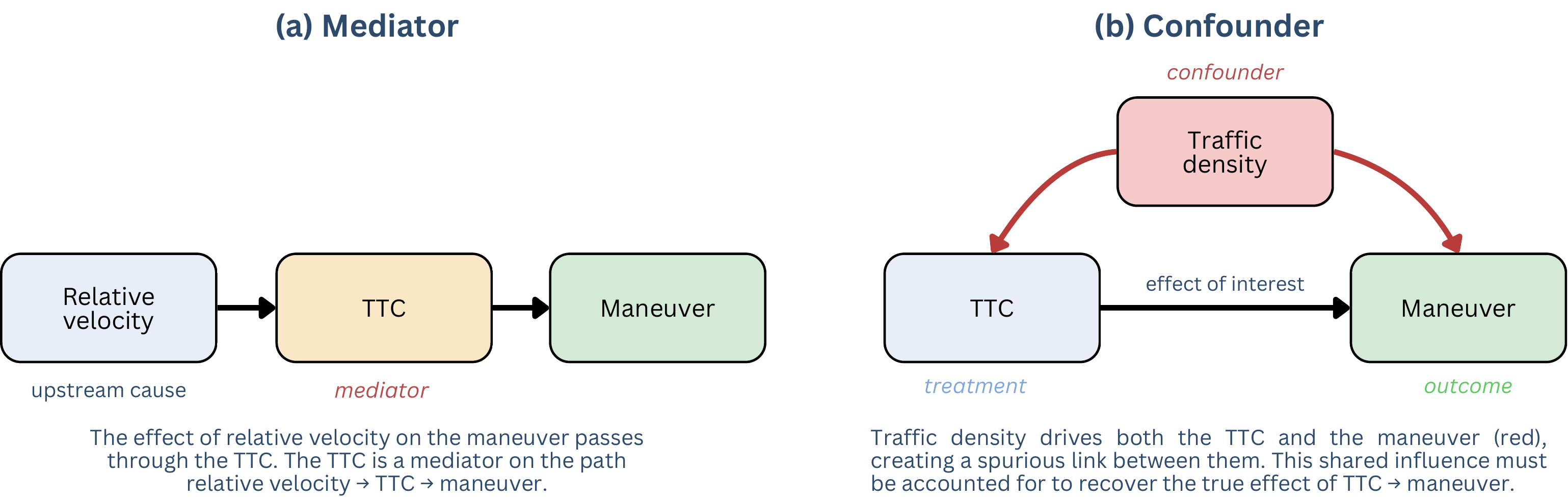}
\caption{Graphical illustration of a mediator and a confounder using representative driving variables. (a) Mediator: the effect of the relative longitudinal velocity on the maneuver passes entirely through the TTC, so the TTC is a mediator lying on the path $\textit{relative velocity} \rightarrow \textit{TTC} \rightarrow \textit{maneuver}$. (b) Confounder: traffic density influences both the TTC and the maneuver (red arrows). Because it drives both of them at the same time, it creates a spurious (non-causal) association between the TTC and the maneuver; this shared influence must be accounted for — by comparing situations at the same traffic-density level — in order to recover the true $\textit{TTC} \rightarrow \textit{maneuver}$ effect shown by the black arrow. These structures motivate the distinction between direct contributors, mediated effects, and confounded relations used throughout the causal analysis.
}
\label{fig:mediator_confounder}
\end{figure}

In the causal-analysis stage, the variable being actively changed is referred to as the \textit{treatment}, while the variable whose response is measured is referred to as the \textit{outcome}. The treatment is not fixed for the whole study; rather, it changes depending on the question being analyzed. For example, when studying the effect of TTC on the maneuver, TTC is the treatment and the maneuver is the outcome. In another analysis, the treatment may be relative velocity and the outcome may be TTC or the maneuver, depending on which downstream effect is being examined. An intervention means setting the treatment to a selected state, such as changing \textit{TTC} from \textit{safe} to \textit{critical}, and then estimating how the selected outcome changes under the learned causal structure. Let $Y$ denote the maneuver outcome and let $X_i$ denote one of the linguistic driving variables. A variable is considered a candidate direct contributor if it has a directed edge toward the maneuver node:
\begin{equation}
X_i \in \mathrm{Pa}_{\mathcal{G}}(Y).
\end{equation}
For example, if TTC has a direct edge toward the maneuver node, then TTC is interpreted as a candidate direct contributor under the learned graph. This does not mean that TTC is always the original reason behind the maneuver; it means that TTC has a direct connection to the maneuver after considering the learned structure. Other variables may influence the maneuver indirectly through a chain of intermediate variables. In this case, the earlier variables are treated as upstream causes, while the intermediate variables are treated as mediators because they transmit the effect toward the maneuver. For example, a path such as
$
\textit{traffic density} \rightarrow \textit{gap} \rightarrow \textit{TTC} \rightarrow Y
$
indicates that traffic density acts upstream, while gap and TTC transmit part of this influence toward the maneuver prediction. The learned graph is also used to identify variables that should be considered during effect estimation. For example, if traffic density affects both TTC and the maneuver, then estimating the effect of TTC on the maneuver without considering traffic density may mix the effect of TTC with the effect of the traffic density. In this case, traffic density acts as a confounder for the TTC--maneuver relation.
To make these roles concrete, \Cref{fig:mediator_confounder} illustrates the two structures using representative driving variables. In \Cref{fig:mediator_confounder}(a), the relative longitudinal velocity influences the maneuver only through the TTC; the TTC is therefore a mediator that transmits the upstream effect along the path
$
\textit{relative velocity} \rightarrow \textit{TTC} \rightarrow \textit{maneuver}.
$
In \Cref{fig:mediator_confounder}(b), traffic density affects both the TTC and the maneuver at the same time. It is therefore a confounder of the TTC–maneuver relation: because it drives both variables, part of the apparent association between the TTC and the maneuver is not produced by the TTC itself, but by their shared dependence on traffic density. To avoid attributing this shared influence to the TTC, traffic density must itself be taken into account when the effect of the TTC on the maneuver is estimated, for example by comparing only situations that share the same traffic-density level. These two diagrams are illustrative examples of the concepts; the dependencies actually learned for this study are reported in the next section.
By distinguishing these roles (direct contributor, mediator, upstream cause, and confounder) the learned graph helps the framework interpret direct effects, mediated effects, and possible shared upstream influences before applying intervention-based analysis. This makes the learned graph more informative than a flat feature-importance ranking: such a ranking may indicate that TTC is important, but it cannot show whether TTC is a direct contributor, a mediator, or a variable affected by upstream traffic conditions. By organizing the variables into these causal roles, the framework can explain not only which variables are related to the predicted maneuver, but also how their effects propagate through the learned causal structure. More details about these graphical structures and their role in causal inference can be found in \cite{pearl2009causal}.

\subsection{Causal Analysis and Explanation}
\label{subsec:causal-analysis}

After learning the \ac{SCM}, the framework uses the trained DECI model for causal analysis and explanation. This stage uses the graph roles introduced in Section~\ref{subsec:causal-roles} to examine how selected treatment changes affect downstream outcomes, whether the estimated effects remain robust under refutation tests, which maneuver is predicted by the trained model, and which causal chain explains why this maneuver is favored over the alternatives. Accordingly, this stage consists of intervention-based effect analysis, refutation testing, maneuver prediction, and contrastive causal-chain explanation. This corresponds to the final causal-analysis block in \Cref{fig:methodology}.

\subsubsection{Intervention-Based Effect Analysis}
\label{subsubsec:intervention-analysis}

Intervention-based effect analysis estimates how the probability of a selected outcome state changes when a treatment variable is actively set to a different linguistic state under the learned causal structure. The maneuver label is the main outcome of interest, but the same causal-effect formulation can also be applied to intermediate child variables in the learned graph. For example, relative velocity can be treated as the treatment when estimating its effect on TTC, while TTC can later be treated as the treatment when estimating its effect on the maneuver. This allows the framework to analyze not only the final maneuver prediction, but also how effects propagate between intermediate variables.

Let $X_i$ denote the selected treatment variable and let $Y$ denote the selected outcome for the current causal query. The \ac{ATE} of changing $X_i$ from state $x$ to state $x'$ on an outcome state $y$ is defined as the change in the interventional probability of that outcome state:
\begin{equation}
\mathrm{ATE}_{x \rightarrow x'}(y) =
P(Y=y \mid do(X_i=x')) - P(Y=y \mid do(X_i=x)).
\end{equation}
Here, $do(X_i=x)$ denotes an intervention that sets the treatment variable $X_i$ to the state $x$ under the learned \ac{SCM}. Since the variables are categorical, the \ac{ATE} is expressed as a difference in probabilities. If the selected outcome $Y$ is the maneuver label, then $y$ can be one of the three maneuver classes: LLC, LK, or RLC. If the selected outcome $Y$ is an intermediate variable, such as TTC, then $y$ can be one of its linguistic states, such as \textit{critical}, \textit{cautious}, or \textit{safe}. In practice, estimating this effect may require accounting for confounders. Since the learned graph contains 47 linguistic driving variables and the maneuver label, a selected treatment may have more than one shared upstream variable that also affects the selected outcome. Manually deciding which variables should be considered for every treatment--outcome pair would be difficult and error-prone. Therefore, DoWhy is used with the learned causal graph to identify the required adjustment set — the set of upstream variables that influence both the treatment and the outcome (the confounders), which must be held at the same level when comparing different treatment states so that their shared influence is not mistaken for the effect of the treatment — and to estimate the intervention effect. The resulting intervention effects are used to distinguish influential treatments from weakly influential treatments for the selected outcome. A treatment is considered influential when changing its state produces a meaningful change in the selected outcome distribution; otherwise, its effect is considered weak or negligible under the learned causal structure.

\subsubsection{Refutation and Robustness Testing}
\label{subsubsec:refutation-testing}

After estimating the intervention-based effects, refutation tests are applied to examine whether the estimated effects behave as expected under controlled diagnostic changes. These tests are not used to prove that the learned causal graph is correct. Instead, they provide additional checks on the reliability of the estimated treatment effect. If an effect behaves unexpectedly under a refutation test, the corresponding treatment--outcome relation should be interpreted with caution and may require further investigation. Refutation testing is performed using DoWhy after the intervention effect has been estimated for a selected treatment--outcome pair. Each test modifies the estimation setting in a different way and checks whether the resulting behavior is reasonable for that test. Therefore, the expected result is not identical for all refutation tests. Three refutation tests are used. The placebo-treatment test replaces the original treatment with an artificial or randomly generated treatment. Since this placebo treatment should not have a meaningful causal effect on the outcome, the estimated effect is expected to become weak or close to zero. If the placebo treatment still produces a strong effect, this suggests that the original treatment--outcome estimate may be influenced by spurious associations in the data. The random-common-cause test adds a randomly generated variable as a simulated additional common cause between the treatment and the outcome. In this case, the estimated effect is expected to remain reasonably close to the original estimate. If adding this simulated common cause substantially changes the estimated effect, this does not mean that the random variable is a real confounder. Rather, it indicates that the estimated effect is sensitive to changes in the adjustment setting, and therefore the corresponding treatment--outcome relation should be interpreted with caution. The data-subset test repeats the effect estimation using a subset of the data. The purpose is to check whether the estimated effect depends strongly on a specific portion of the dataset. If the effect remains similar across subsets, this provides additional support for the robustness of the estimate. If the effect changes strongly, the treatment--outcome relation may be sample-dependent and should be examined further. Overall, the refutation stage provides an additional reliability check for the intervention-based effect analysis. Passing these tests does not prove definitive causality, but it increases confidence that the estimated effect is not simply produced by placebo treatments, random common causes, or a specific subset of the data. Conversely, failing a refutation test does not identify the exact reason for the failure, such as a specific missing confounder or problematic data segment, but it highlights a relation that should be treated cautiously before being used in the causal explanation.

\subsubsection{Maneuver Prediction}
\label{subsubsec:maneuver-prediction}

The trained DECI model is also used to predict the future maneuver for each traffic sample. In this step, the outcome is specifically the maneuver label, denoted by $Y$. Given the linguistic feature states of the considered sample, the model estimates the probability of each maneuver class:
\begin{equation}
P(Y \mid \mathbf{X}) =
\left[
P(Y=\mathrm{LLC}\mid \mathbf{X}),
P(Y=\mathrm{LK}\mid \mathbf{X}),
P(Y=\mathrm{RLC}\mid \mathbf{X})
\right],
\end{equation}
where $\mathbf{X}$ denotes the set of linguistic driving variables for the considered sample. The predicted maneuver is selected as the class with the highest probability:
\begin{equation}
\hat{Y} =
\arg\max_{y \in \{\mathrm{LLC},\mathrm{LK},\mathrm{RLC}\}}
P(Y=y \mid \mathbf{X}).
\end{equation}
The full probability distribution is retained because it provides more information than the final class label alone. It shows not only which maneuver is most likely, but also how strongly the model favors this maneuver compared with the alternatives. This prediction step is performed using the trained DECI-based \ac{SCM}, not a separate flat classifier. The learned causal graph is then used in the following explanation stage, where the current linguistic states of the sample are traced through their learned causal parents to explain why the predicted maneuver is supported.

\subsubsection{Contrastive Causal-Chain Explanation}
\label{subsubsec:causal-chain-explanation}

After predicting the maneuver, the learned graph is used to generate a recursive causal-chain explanation for the predicted class by identifying which observed parent states most strongly support this prediction and which upstream variables explain those parent states. The explanation therefore starts from the maneuver node, ranks its direct parents according to their local interventional effect on the predicted maneuver, and then moves backward through the learned graph to explain the state of the selected parent. Let $X_i$ be a parent of $X_j$ in the learned graph. For a given sample, suppose that the observed state of the parent is $X_i=x_i$ and the observed state of the child is $X_j=x_j$. The local interventional lift from $X_i$ to $X_j$ is defined as
\begin{equation}
\Delta_{i \rightarrow j}(x_j)
=
P(X_j=x_j \mid do(X_i=x_i)) - P(X_j=x_j).
\end{equation}
Here, $P(X_j=x_j)$ denotes the baseline model-implied probability of the observed child state under the learned \ac{SCM}, before imposing the observed parent state through an intervention. The explanation is local because $x_i$ and $x_j$ correspond to the observed parent and child states in the current sample. Therefore, $\Delta_{i \rightarrow j}(x_j)$ measures how much the observed parent state increases or decreases the model base probability of the observed child state. For the final edge ending at the maneuver node, $X_j$ is the maneuver label $Y$. Thus, if the predicted maneuver is $\hat{Y}$, the framework evaluates
\begin{equation}
\Delta_{i \rightarrow Y}(\hat{Y})
=
P(Y=\hat{Y} \mid do(X_i=x_i)) - P(Y=\hat{Y}).
\end{equation}
A positive lift means that the observed parent state supports the predicted maneuver, while a negative lift means that it suppresses that maneuver. For example, assume that the predicted maneuver is left lane change, $\hat{Y}=\mathrm{LLC}$, and that one direct parent of the maneuver is the preceding-vehicle TTC, with the observed state $\mathrm{TTC}_{\mathrm{prec}}=\mathrm{critical}$. If the baseline probability is $P(Y=\mathrm{LLC})=0.24$, and intervening on the observed TTC state gives
\[
P(Y=\mathrm{LLC}\mid do(\mathrm{TTC}_{\mathrm{prec}}=\mathrm{critical}))=0.43,
\]
then the local lift is $0.43-0.24=0.19$. This means that the critical TTC state increases the model-implied probability of LLC by 0.19, so it is treated as a supporting direct contributor to the prediction. The explanation does not stop at this direct parent. It recursively asks why this parent has its observed state. Continuing the same example, suppose that the current sample has $\mathrm{relLongVel}_{\mathrm{prec}}=\mathrm{target\_is\_much\_faster}$ and $\mathrm{TTC}_{\mathrm{prec}}=\mathrm{critical}$. The framework then evaluates the effect of the observed relative-velocity state on the observed TTC state. If
\[
P(\mathrm{TTC}_{\mathrm{prec}}=\mathrm{critical})=0.18
\]
and
\[
P(\mathrm{TTC}_{\mathrm{prec}}=\mathrm{critical}
\mid do(\mathrm{relLongVel}_{\mathrm{prec}}=\mathrm{target\_is\_much\_faster}))=0.62,
\]
then the local lift is $0.62-0.18=0.44$. This indicates that the observed relative velocity is a strong upstream contributor to the critical TTC state. Accordingly, one possible explanation chain for this prediction is
\begin{equation}
\mathrm{relLongVel}_{\mathrm{prec}}=\mathrm{target\_is\_much\_faster}
\rightarrow
\mathrm{TTC}_{\mathrm{prec}}=\mathrm{critical}
\rightarrow
Y=\mathrm{LLC}.
\end{equation}
This path indicates that the model does not explain the LLC by TTC alone. Instead, it traces how the current relative-velocity state supports the critical TTC state, and how the critical TTC state supports the predicted LLC. More generally, a causal explanation path is a directed path in the learned graph that ends at the maneuver node:
\begin{equation}
X_{p_1} \rightarrow X_{p_2} \rightarrow \cdots \rightarrow X_{p_k} \rightarrow Y,
\end{equation}
where each edge is selected because the observed parent state produces a meaningful local interventional lift for the observed child state or for the target maneuver. For a longer interaction sequence, the explanation may take the form
\begin{equation}
\textit{traffic density}
\rightarrow
\textit{gap}
\rightarrow
\textit{TTC}
\rightarrow
\mathrm{LLC}.
\end{equation}
This means that the prediction is interpreted as a mechanism that propagates through upstream traffic and interaction variables, rather than as the isolated effect of one feature. The same mechanism is used for contrastive explanations. For the predicted maneuver, the framework searches for observed parent states with positive lift toward the predicted class. For a ``why not'' question, it searches for observed parent states with negative lift toward the alternative class. For example, if the question is why the vehicle did not perform an RLC, the framework evaluates direct parents of $Y$ that reduce $P(Y=\mathrm{RLC})$. If the current lane-position state is $\mathrm{laneRank}=\mathrm{rightmost\_lane}$, the baseline probability is $P(Y=\mathrm{RLC})=0.28$, and the intervention gives
\[
P(Y=\mathrm{RLC}\mid do(\mathrm{laneRank}=\mathrm{rightmost\_lane}))=0.08,
\]
then the lift is $0.08-0.28=-0.20$. This negative lift means that the current lane position suppresses the probability of RLC, making it part of the explanation for why RLC is not selected. By tracing these paths recursively, the framework provides explanations that are more informative than a single feature-importance value. A feature-importance method may indicate that TTC is important, but it does not show whether TTC is a direct contributor, a mediator, or a consequence of upstream traffic conditions. The proposed causal-chain explanation links the predicted maneuver to the learned DECI graph and to local intervention-based probability shifts, explaining both why the selected maneuver is favored and why the alternative maneuvers are less supported under the learned \ac{SCM}.

\section{Results and Analysis}

\subsection{Learned Causal Structure}
\label{subsec:learned-causal-structure}

\begin{figure}[pos=!htbp]
\centering
\includegraphics[width=0.5\linewidth]{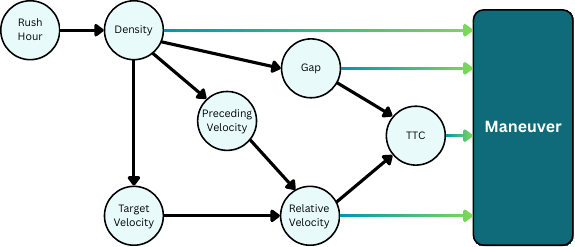}
\caption{Representative subgraph from the learned DECI causal structure. The graph illustrates how contextual variables, such as \textit{rush hour}, affect the maneuver mainly through intermediate variables rather than only through direct connections. The structure also respects the temporal constraint that no causal edges originate from the maneuver node. Interaction variables such as \textit{gap}, \textit{relative velocity}, and \textit{TTC} appear as key downstream variables that can directly support the maneuver prediction, while upstream variables such as \textit{rush hour} and vehicle-speed variables can influence the maneuver indirectly through these mediators.}
\label{fig:learned-structure}
\end{figure}

Before analyzing intervention effects and prediction outcomes, the learned DECI graph is first examined to verify whether the resulting structure is logically consistent with the expert-constrained formulation. The objective of this step is to inspect whether the learned dependencies follow physically and temporally plausible relations among the driving variables. As shown in \Cref{fig:learned-structure}, the learned graph follows the intended causal logic of the problem. First, the maneuver node appears only as a downstream outcome, with no outgoing edges toward the observed traffic variables. This is consistent with the constraint formulation used during training.
The figure also shows that not all variables influence the maneuver in the same way. For example, \textit{rush hour} does not point directly to the maneuver node; instead, it acts as an upstream contextual variable that affects \textit{density}. In contrast, \textit{density} has a direct connection to the maneuver, but it also influences the maneuver indirectly through other interaction variables. For instance, denser traffic may reduce the available gap, affect the target and preceding-vehicle speed states, and contribute to a higher relative velocity when the target vehicle is moving faster than the preceding vehicle. Therefore, density can contribute to the maneuver both directly and through intermediate variables, even if its direct effect may be weaker than more immediate interaction variables. Several intermediate relations learned by the model are also physically meaningful. For example, \textit{density} is connected to \textit{preceding velocity}, \textit{target velocity}, and \textit{gap}, reflecting the idea that denser traffic conditions can reduce available spacing and influence surrounding-vehicle motion. The graph further connects \textit{target velocity} and \textit{preceding velocity} to \textit{relative velocity}, and connects \textit{gap} together with \textit{relative velocity} to \textit{TTC}. These relations are consistent with the physical interpretation of the variables, since TTC depends on both spacing and relative motion between vehicles. The learned structure therefore suggests that the maneuver is not driven by a single isolated feature, but by interacting motion and contextual variables. In some situations, the prediction may be mainly supported by a small gap or a critical TTC state; in other situations, the dominant support may come from relative velocity or from broader contextual conditions such as traffic density. This graph therefore shows the possible paths through which the variables can influence the maneuver. After verifying that the learned structure is logically organized, the following subsection uses intervention analysis to quantify how strongly selected variables affect the maneuver probabilities under the learned causal model.

\subsection{Intervention-Based Effect Analysis}
\label{subsec:intervention-results}

\begin{figure}[pos=!htbp]
\centering
\includegraphics[width=0.8\linewidth]{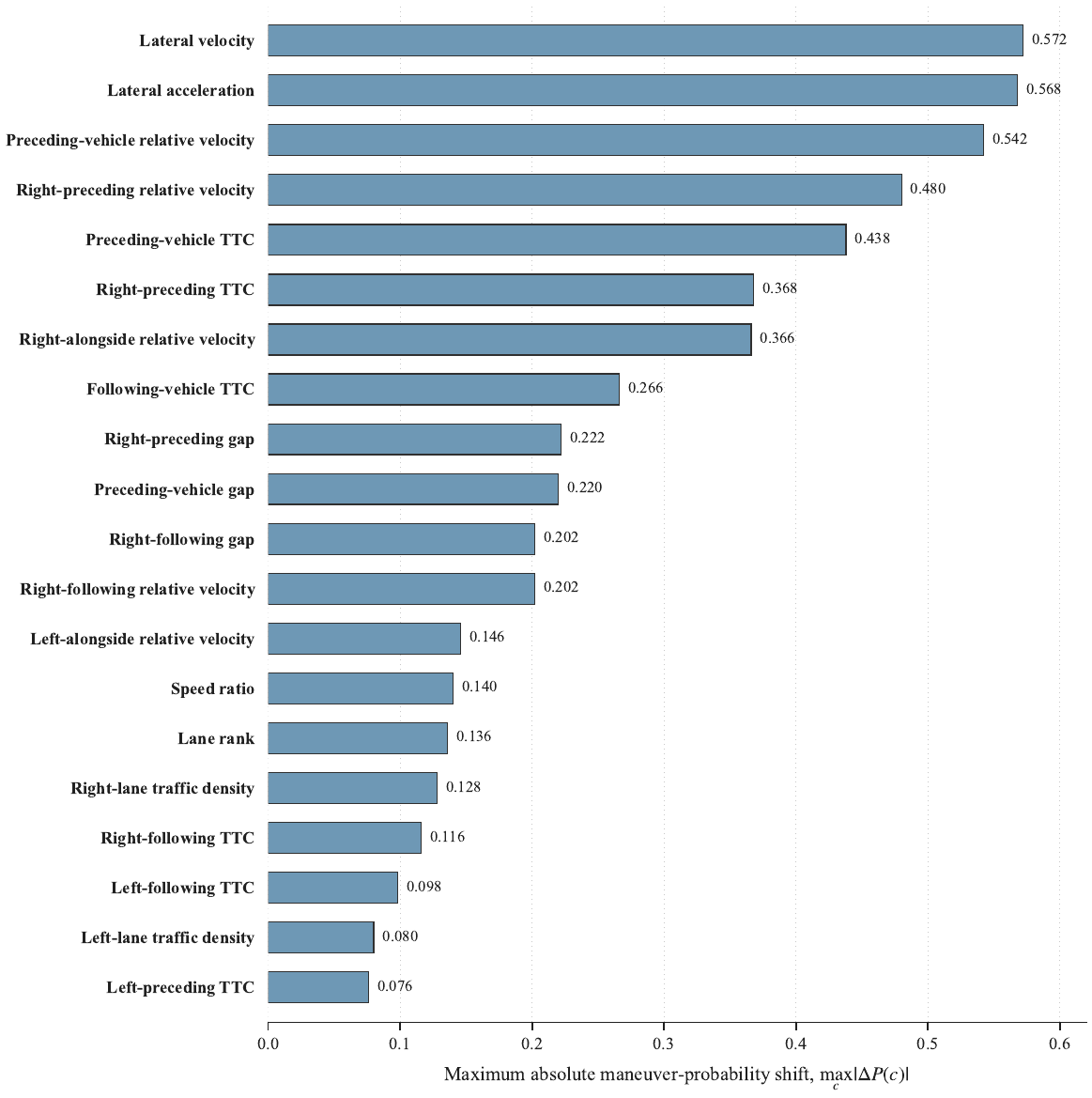}
\caption{Intervention-effect magnitude ranking of the driving variables under the learned DECI-based \ac{SCM}. Each bar shows the maximum absolute maneuver-probability shift produced by intervening on a feature state, computed across the three maneuver classes. Larger values indicate that changing the corresponding feature can produce a stronger change in at least one maneuver probability.}
\label{fig:intervention-effect-ranking}
\end{figure}

\begin{figure*}[pos=!htbp]
\centering
\includegraphics[width=\textwidth]{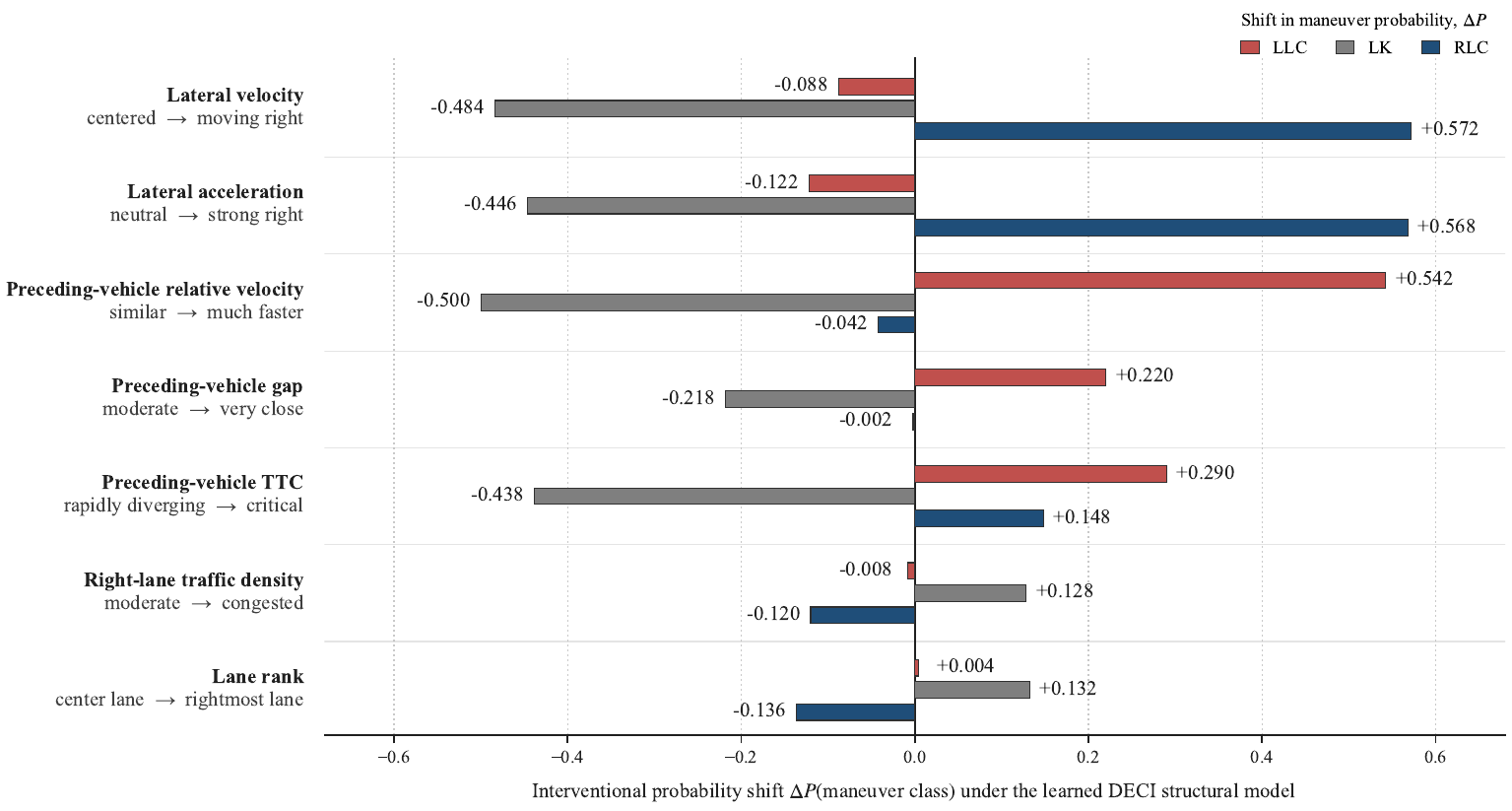}
\caption{Signed class-specific intervention effects for representative feature-state changes. Each row shows the change in the probabilities of LLC, LK, and RLC after intervening on a selected feature state under the learned DECI-based \ac{SCM}. Positive values indicate that the intervention increases the probability of the corresponding maneuver class, while negative values indicate that the intervention decreases it. Since LLC, LK, and RLC form one probability distribution, the class-specific shifts for each intervention sum approximately to zero.}
\label{fig:intervention-effect-signed-shifts}
\end{figure*}

After inspecting the learned causal structure, intervention-based effect analysis is used to quantify how selected feature-state changes affect the maneuver probabilities under the learned DECI-based \ac{SCM}. \Cref{fig:intervention-effect-ranking} ranks the variables according to their largest absolute class-specific probability shift.
The strongest effects are obtained for lateral-motion variables. This is expected because these variables directly describe the lateral movement of the target vehicle, which is highly informative for distinguishing lane keeping from left or right lane changes. However, the ranking also shows that the model does not rely only on lateral motion. Interaction variables, including preceding-vehicle relative velocity, right-preceding relative velocity, TTC, and gap-related variables, also produce large shifts in the maneuver distribution. This indicates that the learned \ac{SCM} captures both the lateral motion of the target vehicle and the surrounding-vehicle interactions that may motivate a lane-change maneuver. Contextual variables, such as lane rank, speed ratio, and lane-level traffic density, produce smaller shifts compared with the strongest lateral-motion and interaction variables. This is consistent with the interpretation of the learned graph: contextual variables may contribute to the maneuver directly, but they can also affect the maneuver indirectly through downstream interaction variables such as gap, relative velocity, and TTC. Therefore, a smaller direct intervention effect does not necessarily mean that the variable is irrelevant; because maybe the variable influence is partially transmitted through other intermediate variables (mediators). \Cref{fig:intervention-effect-signed-shifts} provides a more detailed view of the intervention results by showing the signed probability shifts for representative feature-state changes. This figure is important because it allows the direction of the learned effects to be checked, not only their magnitude. For example, changing lateral velocity from \textit{centered} to \textit{moving\_right} increases the probability of RLC and decreases the probabilities of the other maneuver classes. Similarly, changing lateral acceleration from \textit{neutral} to \textit{strong\_right} produces a strong positive shift toward RLC. These effects are consistent with the expected interpretation of rightward lateral motion.
The same logic appears for interaction-related variables. When the preceding-vehicle relative velocity changes from \textit{similar} to \textit{much\_faster}, the probability of LLC increases. This reflects a situation in which the target vehicle is moving much faster than the preceding vehicle, creating pressure to leave the current lane. A similar interpretation applies when the preceding gap changes from \textit{moderate} to \textit{very\_close}; the reduced spacing increases the support for a lane-change maneuver. Similarly, when the preceding TTC changes from \textit{diverging\_fast} to \textit{critical}, the probability of lane keeping decreases, indicating that a critical interaction with the preceding vehicle makes maintaining the current lane less supported by the model.
The figure also shows that contextual interventions behave consistently with driving constraints. For example, changing lane rank from \textit{center\_lane} to \textit{rightmost\_lane} decreases the probability of RLC. This is logical because a right lane change becomes less feasible or less supported when the vehicle is already in the rightmost lane. The probability mass that is lost from RLC is redistributed to the remaining maneuver classes. This redistribution is expected because the three maneuver probabilities must remain normalized; therefore, the sum of the class-specific probability changes for each intervention is approximately zero. Overall, the intervention results support the causal interpretation of the learned structure. The ranking in \Cref{fig:intervention-effect-ranking} identifies which variables can produce the strongest maneuver-probability changes, while \Cref{fig:intervention-effect-signed-shifts} shows whether these changes follow a logical direction. Together, the figures show that the learned DECI-based \ac{SCM} does not only associate variables with the maneuver label, but also produces meaningful changes in the maneuver distribution when feature states are intervened upon. The following subsection therefore evaluates whether these estimated effects remain stable under refutation and robustness tests.

\subsection{Refutation and Robustness Testing}
\label{subsec:refutation-results}

After estimating the intervention effects, refutation tests are applied to examine whether selected treatment--outcome effects remain reliable under controlled diagnostic changes. This step is important because an intervention effect should not be used for causal interpretation if it appears only because of a random treatment, a specific data subset, or a fragile adjustment setting. In a real driving application, refutation tests are not intended to be executed online for every vehicle sample. Rather, they provide an offline reliability check before using the estimated effects to support causal explanations. If an effect remains stable under these diagnostic tests, it can be interpreted with higher confidence; if it changes strongly, the corresponding treatment--outcome relation should be treated cautiously. The refutation analysis is performed using DoWhy on representative intervention effects obtained from the learned DECI-based \ac{SCM}. Unlike the previous intervention analysis, which reports class-specific shifts for LLC, LK, and RLC, the refutation analysis is applied here to a binary lane-change outcome. Specifically, LLC and RLC are grouped into one lane-change class, denoted as LC, while LK is treated as the non-lane-change class. This formulation is used because the applied DoWhy refutation diagnostics support binary outcome settings only. Therefore, the reported effects represent changes in the probability of performing a lane change, $P(\mathrm{LC})=P(\mathrm{LLC})+P(\mathrm{RLC})$, rather than separate shifts for the three maneuver classes. Three refutation tests are considered. The placebo-treatment test replaces the original treatment with a randomly generated placebo treatment; therefore, the resulting effect is expected to be close to zero. The random-common-cause test adds a simulated common cause to the adjustment setting; in this case, the estimated effect should remain close to the original estimate if the result is robust. The data-subset test repeats the estimation on an 80\% subset of the data; a stable effect should remain close to the original estimate. Table~\ref{tab:refutation-results} summarizes representative refutation results for selected treatment contrasts.
The original effects in \Cref{tab:refutation-results} are consistent with expected driving behavior. For example, changing \textit{Lateral velocity} from \textit{centered} to \textit{moving right} increases the lane-change probability by approximately 0.498. Similarly, changing \textit{Lateral acceleration} from \textit{neutral} to \textit{strong right} increases the lane-change probability by approximately 0.488. These effects are logical because rightward lateral motion indicates that the vehicle is no longer simply maintaining its lane.
\begin{table}[pos=!htbp]
\centering
\caption{Representative DoWhy refutation results for selected intervention effects. The outcome is the binary lane-change probability, $P(\mathrm{LC})=P(\mathrm{LLC})+P(\mathrm{RLC})$, where LLC and RLC are grouped as lane-change maneuvers and LK is treated as the non-lane-change class. The reported values are probability changes, not raw probabilities.}
\label{tab:refutation-results}
\begin{tabular}{p{4.4cm} p{3.7cm} p{1.5cm} p{1.5cm} p{1.5cm} p{1.3cm}}
\hline
\textbf{Treatment} & \textbf{Contrast} & \textbf{Original effect} & \textbf{Placebo effect} & \textbf{Random common cause} & \textbf{Data subset} \\
\hline
\textit{Lateral velocity}
& \textit{centered} $\rightarrow$ \textit{moving right}
& 0.498
& -0.001
& 0.501
& 0.505 \\

\textit{Lateral acceleration}
& \textit{neutral} $\rightarrow$ \textit{strong right}
& 0.488
& -0.002
& 0.482
& 0.486 \\

\textit{Preceding TTC}
& \textit{diverging fast} $\rightarrow$ \textit{critical}
& 0.451
& 0.002
& 0.455
& 0.445 \\

\textit{Right-alongside relative velocity}
& \textit{much faster} $\rightarrow$ \textit{similar}
& -0.452
& -0.001
& -0.458
& -0.455 \\

\textit{Preceding relative velocity}
& \textit{similar} $\rightarrow$ \textit{much faster}
& 0.379
& -0.000
& 0.377
& 0.370 \\
\hline
\end{tabular}
\end{table}
The \textit{Preceding TTC} contrast also has a strong positive effect: changing it from \textit{diverging fast} to \textit{critical} increases the lane-change probability by approximately 0.451, which is consistent with the idea that a critical interaction with the preceding vehicle makes a lane change more likely. In contrast, changing \textit{Right-alongside relative velocity} from \textit{much faster} to \textit{similar} decreases the lane-change probability, suggesting that the corresponding right-side interaction becomes less supportive of a lane-change maneuver.

The placebo-treatment effects are close to zero for all reported treatment contrasts. This behavior is expected because the placebo treatment is randomly generated and should not explain the lane-change outcome. Therefore, the near-zero placebo effects indicate that the original estimates are not reproduced when the real treatment variable is replaced by an artificial one. The random-common-cause and data-subset tests also show stable behavior for the reported examples. For instance, the effect of changing \textit{Lateral velocity} from \textit{centered} to \textit{moving right} remains close to 0.50 after adding a random common cause and after using an 80\% subset of the data. Similarly, changing \textit{Preceding TTC} from \textit{diverging fast} to \textit{critical} produces an original effect of 0.451, while the random-common-cause and subset estimates remain close to this value. These results indicate that the selected effects are not strongly altered by the diagnostic perturbations. 
It is important to note that passing the refutation tests does not prove definitive real-world causality. Rather, it provides additional support that the reported intervention effects are stable under the tested perturbations. Conversely, effects that produce numerically unstable estimates should be treated cautiously and are not used as primary evidence in the causal explanation. Overall, the refutation results support the reliability of the main intervention effects and the model in order to be used in the subsequent prediction and explanation analysis.

\subsection{Maneuver Prediction Performance}
\label{subsec:prediction-results}

\begin{figure}[pos=!htbp]
\centering
\includegraphics[width=\linewidth]{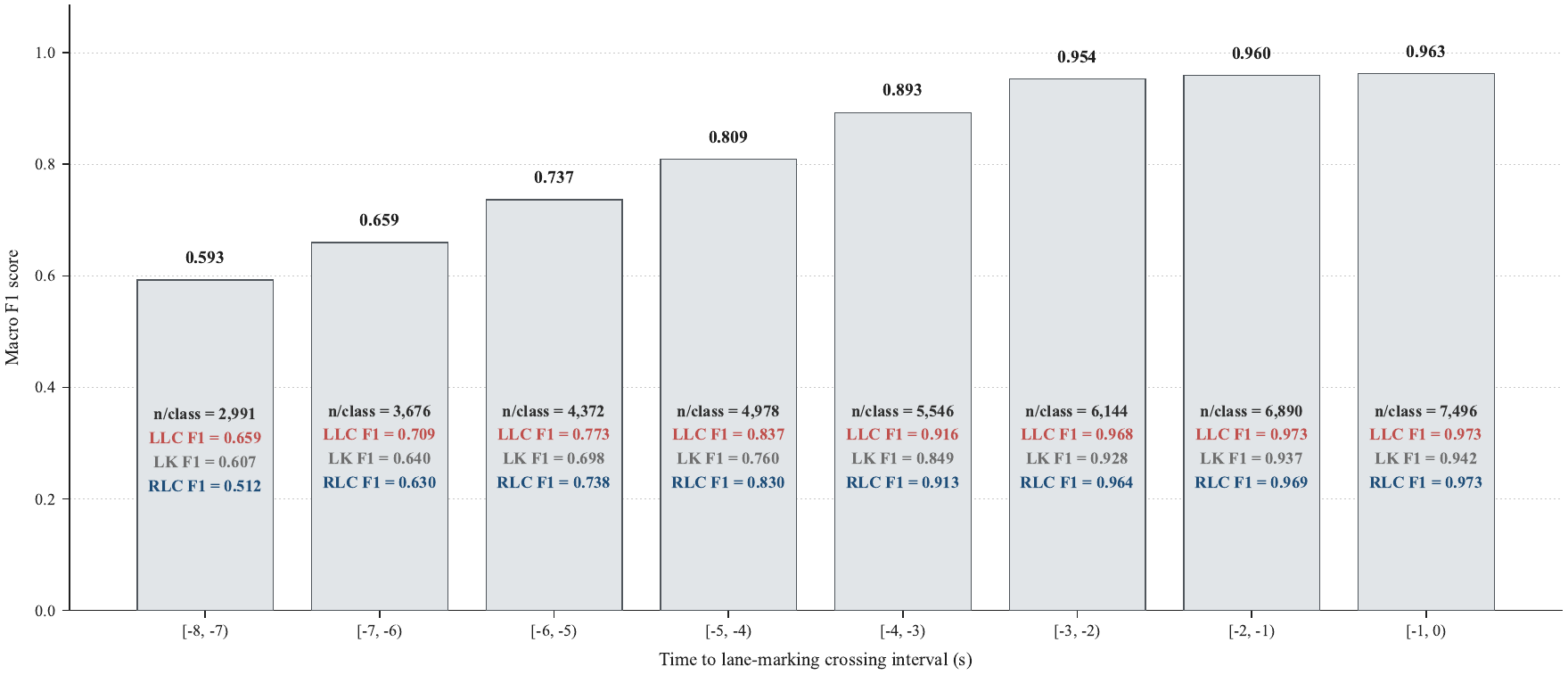}
\caption{Macro F1-score of the DECI-based \ac{SCM} across time intervals before the lane-marking crossing event. The crossing event is defined as $t=0$, and negative intervals indicate time before this event. For example, $[-1,0)$ represents the last second before lane-marking crossing, while $[-8,-7)$ represents the interval from 8 to 7 s before the crossing. The main prediction horizon considered in the experiment is the 5 s interval before the crossing event.}
\label{fig:deci-performance-by-crossing-interval}
\vspace{-3mm}
\end{figure}

After evaluating the intervention effects and their robustness, the trained DECI-based \ac{SCM} is used to predict the maneuver class for each test sample. The objective of this evaluation is to verify that the same structural model used for causal analysis can also provide reliable maneuver probabilities for LLC, LK, and RLC. As shown in \Cref{fig:deci-performance-by-crossing-interval}, prediction performance increases as the target vehicle approaches the lane-marking crossing event. The model achieves strong macro F1-scores in the intervals closest to the event, reaching 96.3\%, 96.0\%, and 95.4\% in the first three seconds before crossing. The score then decreases to 89.3\% in the $(3,4]$ s interval and 80.9\% in the $(4,5]$ s interval. This shows that the model remains effective within the main 5 s prediction horizon, while the additional diagnostic intervals beyond 5 s show the expected performance decrease at earlier anticipation times.
This trend is reasonable because maneuver-related cues are less explicit far from the crossing event. At earlier intervals, the feature patterns of different maneuver classes may overlap, making the future maneuver harder to distinguish.
As $t=0$ becomes closer, lateral-motion variables and interaction-related variables, such as gap, relative velocity, and TTC, become more informative. This behavior is also consistent with the intervention-analysis results, where lateral-motion and interaction-risk variables produced strong shifts in maneuver probabilities.

To position the proposed framework with respect to previous lane-change prediction studies, \Cref{tab:comparison-highd} compares reported F1-scores from representative works that also used the highD dataset. The comparison shows that existing highD-based lane-change prediction models achieve strong F1-scores, especially close to the lane-change event. For example, \cite{xue2022integrated} and \cite{gao2023dual} report high performance up to 2 s before the event. The dual-transformer-based approach in \cite{gao2023dual} reaches very high scores at 0.5 and 1 s, while its score decreases at 2 s. The knowledge-graph-based approach in \cite{manzour2024vehicle} reports results up to 4 s before the event, where the performance also decreases as the prediction time becomes earlier. The proposed framework follows the same general behavior while reporting performance up to 8 s.
\begin{table}[pos=!htbp]
\centering
\caption{Performance comparison of lane-change prediction models on the highD dataset using the F1-score metric. The upper part reports studies using fixed prediction times before the lane-change event, while the lower part reports studies using prediction intervals before the event.}
\label{tab:comparison-highd}
\setlength{\tabcolsep}{4pt}
\begin{tabular}{|c|c|c|c|c|c|c|c|c|}
\hline
Pred. time (s) & 0.5 & 1 & 1.5 & 2 & 2.5 & 3 & 3.5 & 4 \\
\hline
\cite{xue2022integrated}
& 98.2 & 97.1 & 96.6 & 95.2 & -- & -- & -- & -- \\
\hline
\cite{gao2023dual}
& 99.2 & 99.0 & 97.6 & 91.8 & -- & -- & -- & -- \\
\hline
\cite{manzour2024vehicle}
& 97.7 & 97.9 & 98.1 & 98.0 & 97.2 & 93.6 & 82.8 & 66.5 \\
\hline
\hline
Interval (s) & [0,1] & (1,2] & (2,3] & (3,4] & (4,5] & (5,6] & (6,7] & (7,8] \\
\hline
\cite{peng2025lc}
& 98.5 & 98.9 & 98.1 & 93.0 & -- & -- & -- & -- \\
\hline
\cite{manzour2025explainable}
& 95.5 & 95.8 & 92.4 & 76.0 & -- & -- & -- & -- \\
\hline
Proposed framework
& 96.3 & 96.0 & 95.4 & 89.3 & 80.9 & 73.7 & 65.9 & 59.3 \\
\hline
\end{tabular}
\end{table}
Most of the compared studies rely on numerical trajectory and interaction features for maneuver prediction. The LLM-based work in \cite{peng2025lc} reports strong performance and provides textual explanations, but the explanation is generated from the provided input and the model's learned language knowledge rather than from an explicit causal structure. In other words, it does not estimate intervention effects, perform causal refutation tests, or identify causal paths among the driving variables. The works in \cite{manzour2024vehicle} and \cite{manzour2025explainable} move toward a more interpretable representation by converting driving variables into linguistic states. However, these linguistic variables are still mainly used as a flat input feature vector, without explicitly modeling the dependencies among them. In contrast, the proposed framework learns an \ac{SCM} over the linguistic variables. This allows the relations among motion, interaction, contextual, and maneuver variables to be represented as a graph. Therefore, the contribution is not limited to the reported F1-scores. The same trained model is used for maneuver prediction, causal-effect analysis, refutation testing, and contrastive causal-chain explanation. The prediction results show that the model can produce reliable maneuver probabilities, while the previous intervention and refutation results examine whether the learned feature effects are meaningful and stable under diagnostic perturbations.

\subsection{Diagnostic Validation of Intervention-Based Feature Relevance}
\label{subsec:straight-motion-diagnostic}

The previous results showed that intervention analysis can identify variables with strong or weak effects on the maneuver probabilities. To further examine whether this interpretation is meaningful, an additional diagnostic experiment is conducted. The objective is to test whether variables that are estimated to have weak intervention effects can be removed without noticeably degrading the prediction performance. For this purpose, the temporal reference is shifted from the lane-marking crossing event to the moment before the vehicle develops lateral motion. In this setting, the model is trained using the 5 s interval before this new reference point, so the analysis focuses on the period before the lateral action becomes visible. First, a DECI-based model is trained using the full feature set, including \textit{lateral velocity} and \textit{lateral acceleration}. The intervention analysis on this model indicates that these two lateral-motion variables have weak effects on the maneuver label, with \acp{ATE} approximately in the range of 0.07--0.09. This is much lower than the effects observed in the lane-marking-crossing reference setting, where the same variables produced effects around 0.4--0.6. Based on this intervention result, a second DECI-based model is trained on the same data after removing \textit{lateral velocity} and \textit{lateral acceleration} from the input feature set. If the intervention analysis is meaningful, removing variables with weak estimated effects should not cause a clear reduction in prediction performance. Therefore, the comparison between the two models provides a prediction-side validation of the causal-effect interpretation. As shown in \Cref{tab:straight-motion-ablation}, the prediction performance remains very similar after removing \textit{lateral velocity} and \textit{lateral acceleration}. For example, in the last second before the onset reference event, the macro F1-score changes only from 78.5\% to 78.1\%. Across the remaining intervals, the two configurations also remain close to each other, with no consistent degradation after removing the lateral-motion variables. These results support the output of the intervention analysis. Since the causal-effect analysis estimated weak effects for \textit{lateral velocity} and \textit{lateral acceleration} in this setting, their removal was expected to have limited influence on the prediction results. The ablation experiment confirms this expectation: variables that were estimated to have weak effects also had little impact when removed from the model input.
\begin{table}[pos=!htbp]
\centering
\caption{Diagnostic prediction results before the onset of lateral motion. The table compares macro F1-scores with and without lateral velocity and lateral acceleration. Values are reported as percentages. The interval $[0,1]$ denotes the last second before the onset reference event, while larger intervals denote earlier anticipation windows.}
\label{tab:straight-motion-ablation}
\setlength{\tabcolsep}{4pt}
\begin{tabular}{|c|c|c|c|c|c|c|c|c|}
\hline
Interval (s) & [0,1] & (1,2] & (2,3] & (3,4] & (4,5] & (5,6] & (6,7] & (7,8] \\
\hline
With lateral features
& 78.5 & 78.1 & 77.9 & 78.3 & 78.0 & 76.9 & 74.4 & 71.6 \\
\hline
Without lateral features
& 78.1 & 79.2 & 77.9 & 77.9 & 76.9 & 76.5 & 71.8 & 72.4 \\
\hline
\end{tabular}
\vspace{-3mm}
\end{table}
This diagnostic study demonstrates the practical value of the intervention-based analysis. A conventional prediction evaluation would only report the final F1-score, but it would not indicate whether each input variable is actually influential in the considered setting. In contrast, the proposed causal analysis can identify variables with strong or weak effects, and the ablation result shows that this information is reflected in the prediction behavior of the model. Therefore, the intervention analysis is not only an explanatory component; it can also guide feature interpretation and help identify variables that may be ignored without significantly affecting prediction performance.

\subsection{Contrastive Causal-Chain Explanation Results}
\label{subsec:causal-chain-results}

This subsection applies the contrastive causal-chain explanation procedure introduced in \Cref{subsubsec:causal-chain-explanation} to individual driving sequences. In that procedure, the explanation starts from the predicted maneuver, identifies observed parent states that support this prediction through positive local intervention effects, and recursively traces upstream variables that explain these parent states. For the alternative maneuver classes, the same mechanism is used to identify observed states that reduce their support. Therefore, the qualitative examples in this subsection do not only show the predicted maneuver over time; they also show why the selected maneuver is favored and why the alternative maneuvers are less supported under the learned DECI-based \ac{SCM}. For each example, $t=0$ denotes the lane-marking crossing event, and negative times indicate seconds before this event. The displayed frames therefore illustrate how the model output evolves before the maneuver is visually completed. In all displayed frames, the white vehicle denotes the target vehicle whose future maneuver is being predicted, while the arrow overlaid on the target vehicle indicates the currently predicted maneuver direction. Two examples are considered: one right lane change (RLC) and one left lane change (LLC). The full temporal evolution of both examples is provided as supplementary videos.

\begin{figure}[pos=!htbp]
\centering
\begin{subfigure}[t]{0.32\textwidth}
\centering
\includegraphics[width=\linewidth]{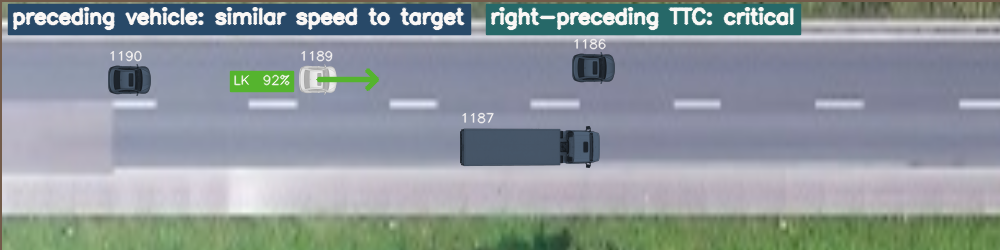}
\caption{$t=-7$ s. The target vehicle is still predicted as LK with high confidence. The overlay indicates that the preceding vehicle has a similar speed to the target, while the right-preceding TTC is critical; however, the evidence is not yet sufficient to favor RLC.}
\label{subfig:rlc-frame-minus7}
\end{subfigure}
\hfill
\begin{subfigure}[t]{0.32\textwidth}
\centering
\includegraphics[width=\linewidth]{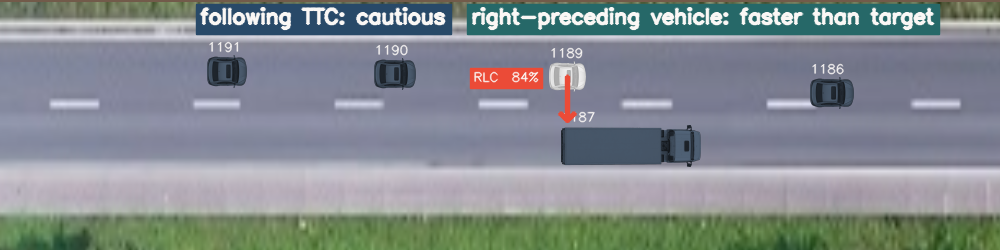}
\caption{$t=-4$ s. The model predicts RLC with 84\% confidence, shortly after the first RLC prediction at $t=-4.52$ s. The displayed cues indicate cautious following TTC and a right-preceding vehicle moving faster than the target, supporting a safe rightward maneuver.}
\label{subfig:rlc-frame-minus4}
\end{subfigure}
\hfill
\begin{subfigure}[t]{0.32\textwidth}
\centering
\includegraphics[width=\linewidth]{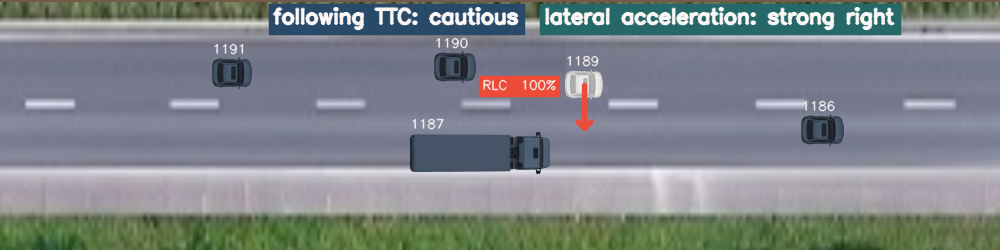}
\caption{$t=-1$ s. The RLC prediction reaches 100\% confidence as the maneuver becomes visually clear. The overlay shows strong rightward lateral acceleration together with cautious following TTC, indicating that the vehicle is already executing the right lane change.}
\label{subfig:rlc-frame-minus1}
\end{subfigure}
\caption{Representative frames from the RLC sequence. The model initially favors LK at $t=-7$ s, predicts RLC for the first time at $t=-4.52$ s, and reaches a confident RLC prediction before the lane-marking crossing event. The sequence illustrates that the maneuver is anticipated before it is visually completed.}
\label{fig:qualitative-rlc-frames}
\vspace{-3mm}
\end{figure}

\begin{figure}[pos=!htbp]
\centering
\includegraphics[width=\textwidth]{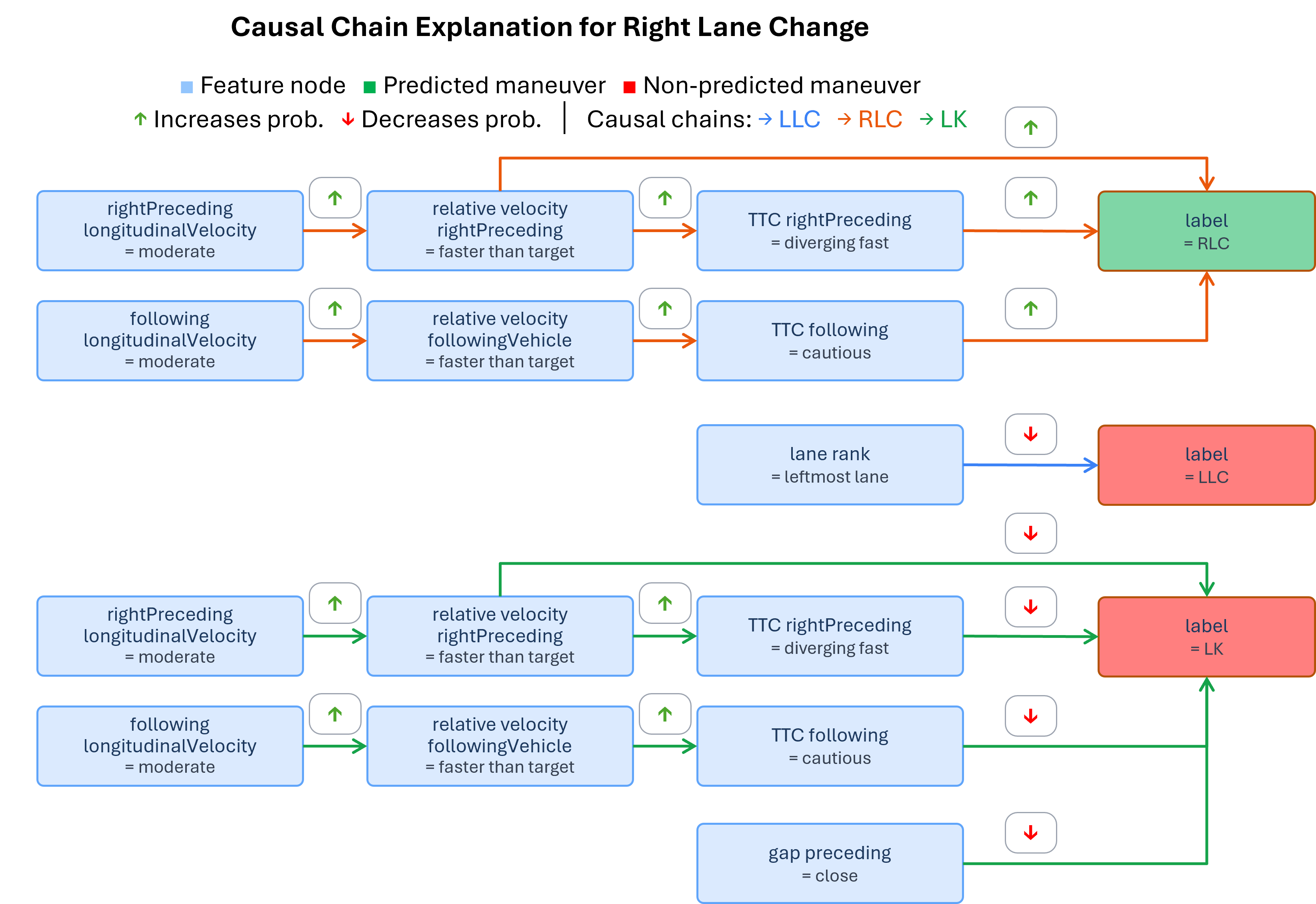}
\caption{Contrastive causal-chain explanation for the RLC example. The orange paths support the predicted RLC maneuver. The right-preceding and following vehicle speed states influence their corresponding relative-velocity states, which then influence the right-preceding and following TTC. The blue path shows that the lane-rank state decreases support for LLC, while the green paths show that the current interaction states reduce support for LK.}
\label{fig:qualitative-rlc-causal-chain}
\vspace{-3mm}
\end{figure}

\Cref{fig:qualitative-rlc-frames} shows the temporal evolution of the prediction in the RLC example. In \Cref{subfig:rlc-frame-minus7} at $t=-7$ s, the target vehicle is still mainly interpreted as lane keeping. Around $t=-4.52$ s, the prediction changes to RLC. Then, the frame at $t=-4$ s shows an RLC probability of 84\%. By $t=-1$ s, the target vehicle is already performing the maneuver and the RLC probability reaches 100\%. This example therefore shows that the model does not only recognize the maneuver after it becomes obvious, but anticipates it several seconds before lateral motion development.
The causal-chain explanation for the RLC example in \Cref{subfig:rlc-frame-minus4} is shown in \Cref{fig:qualitative-rlc-causal-chain}. Following the explanation procedure in \Cref{subsubsec:causal-chain-explanation}, the graph is interpreted from the maneuver node backward through the observed parent states that support the prediction. The orange paths explain why RLC is favored. One branch starts from the right-preceding vehicle longitudinal velocity, which supports the observed relative velocity with the right-preceding vehicle. This relative-velocity state then supports the right-preceding TTC state, which contributes positively to the RLC label. A second branch follows the same logic through the following vehicle: the following-vehicle longitudinal velocity supports the relative velocity with the following vehicle, which then supports the following TTC state and contributes to RLC.
The contrastive part of the same explanation shows why the alternative maneuver classes are less supported. The blue branch indicates that the lane-rank state reduces support for LLC. This is consistent with the scene because the current lane context makes a left lane change infeasible or less supported. The green branches show that the current interaction states, including the right-preceding TTC, following TTC, and preceding gap, reduce support for LK. Thus, the explanation is not limited to saying that RLC has the highest probability; it also identifies why LLC and LK are less consistent with the observed states in this sample. 

\begin{figure}[pos=!htbp]
\centering
\begin{subfigure}[t]{0.32\textwidth}
\centering
\includegraphics[width=\linewidth]{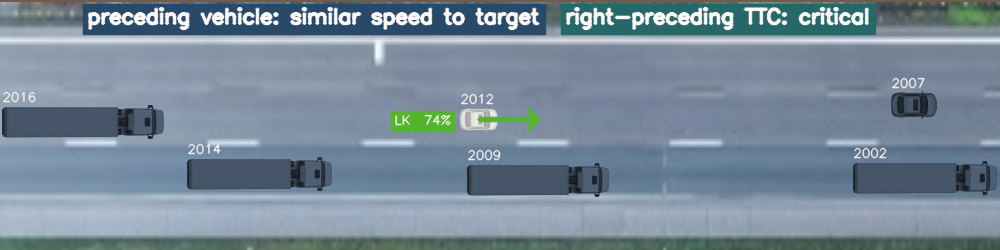}
\caption{$t=-8$ s. The model favors LK with 74\% confidence. The overlay shows a preceding vehicle with similar speed and a critical right-preceding TTC, but the left-lane-change evidence has not yet become dominant.}
\label{subfig:llc-frame-minus8}
\end{subfigure}
\hfill
\begin{subfigure}[t]{0.32\textwidth}
\centering
\includegraphics[width=\linewidth]{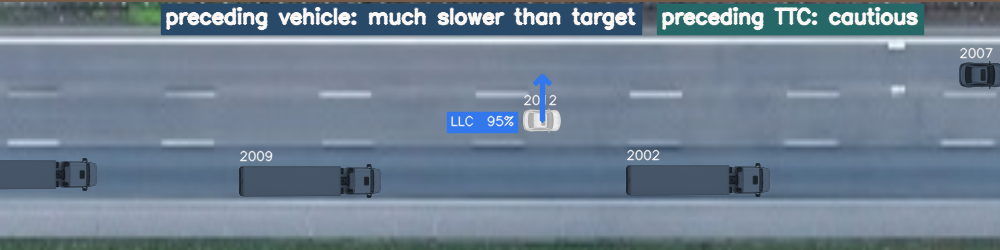}
\caption{$t=-5$ s. The model predicts LLC with 95\% confidence. The first LLC prediction occurs earlier, at $t=-7.16$ s. The overlay indicates that the preceding vehicle is much slower than the target and that the preceding TTC is cautious, supporting a leftward maneuver.}
\label{subfig:llc-frame-minus5}
\end{subfigure}
\hfill
\begin{subfigure}[t]{0.32\textwidth}
\centering
\includegraphics[width=\linewidth]{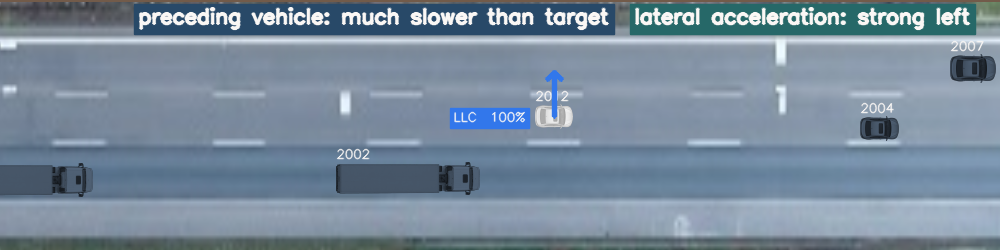}
\caption{$t=-2$ s. The LLC prediction reaches 100\% confidence as the maneuver becomes visually clear. The overlay shows a much slower preceding vehicle and strong leftward lateral acceleration, indicating that the vehicle is already executing the left lane change.}
\label{subfig:llc-frame-minus2}
\end{subfigure}
\caption{Representative frames from the LLC sequence. The model initially favors LK at $t=-8$ s, predicts LLC for the first time at $t=-7.16$ s, and maintains a confident LLC prediction before the lane-marking crossing event.}
\label{fig:qualitative-llc-frames}
\vspace{-3mm}
\end{figure}

\begin{figure}[pos=!htbp]
\centering
\includegraphics[width=\textwidth]{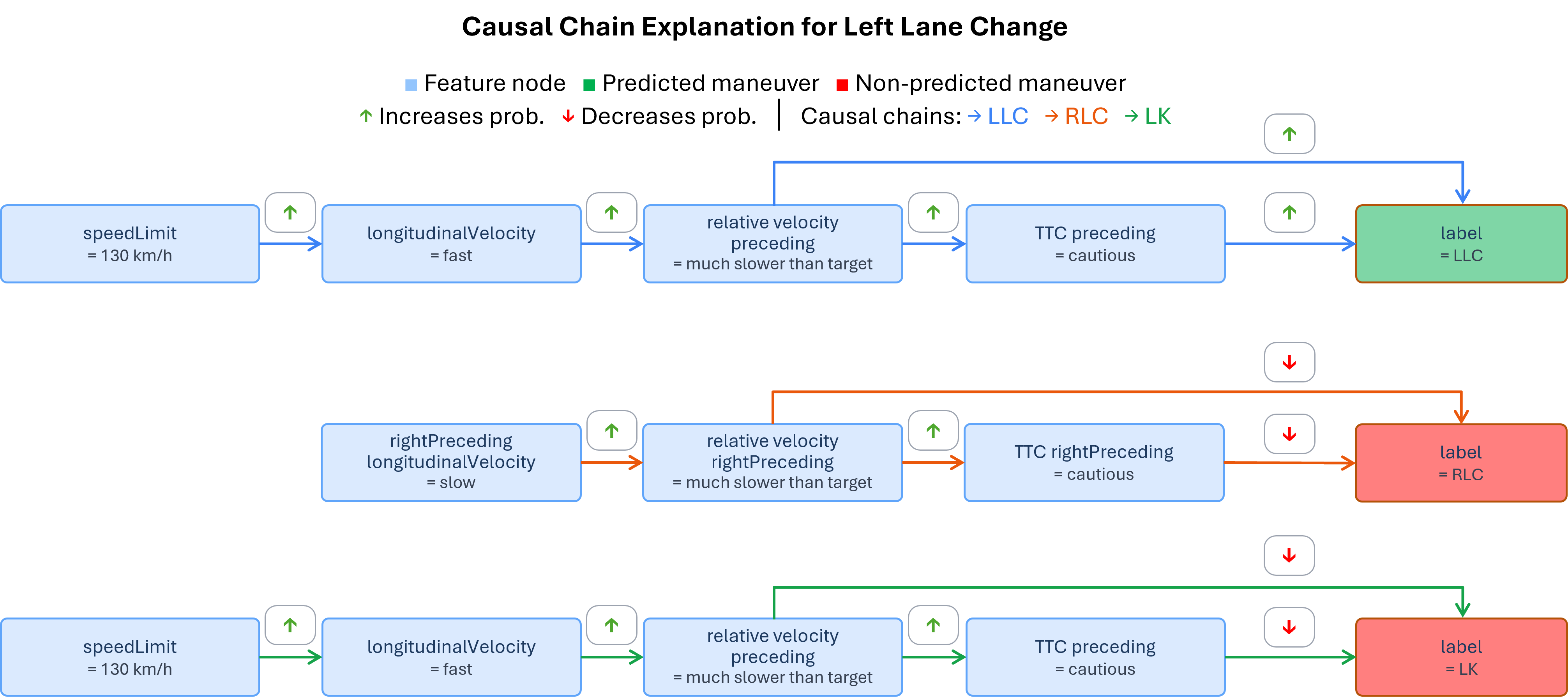}
\caption{Contrastive causal-chain explanation for the LLC example. The blue path supports the predicted LLC maneuver. The speed-limit state supports a fast target-vehicle longitudinal velocity, which contributes to a relative velocity state where the preceding vehicle is much slower than the target. This state affects the preceding TTC and supports LLC. The orange and green paths show states that reduce support for RLC and LK, respectively.}
\label{fig:qualitative-llc-causal-chain}
\vspace{-5mm}
\end{figure}

\Cref{fig:qualitative-llc-frames} presents the temporal evolution of the prediction in the LLC example. At $t=-8$ s, the model still assigns the highest probability to LK, indicating that the lane-change intention is not yet dominant in the visible frame. Shortly after this time, the model first predicts LLC at $t=-7.16$ s. By $t=-5$ s, the LLC probability has increased to 95\%, supported by the interaction with a slower preceding vehicle. At $t=-2$ s, the target vehicle is clearly moving left and the model reaches 100\% confidence for LLC.
The causal-chain explanation for the LLC example in \Cref{subfig:llc-frame-minus5} is shown in \Cref{fig:qualitative-llc-causal-chain}. The blue path explains why LLC is supported in this sample. The chain starts from the speed-limit context, which supports a fast longitudinal velocity for the target vehicle. This target-vehicle velocity contributes to the relative-velocity state with the preceding vehicle, where the preceding vehicle is much slower than the target. This interaction then affects the preceding TTC state, and the resulting chain supports the LLC label. Therefore, the prediction is not explained by a single isolated feature, but by a connected sequence of contextual, kinematic, and interaction-related states.
The contrastive branches explain why the alternatives are less supported. The orange branch indicates that the right-preceding interaction path reduces support for RLC. The green branch indicates that the same upstream context and preceding-vehicle interaction reduce support for LK. This is consistent with the situation in which the target vehicle is fast while the preceding vehicle is much slower, making lane keeping less supported than a left lane change. Overall, these qualitative examples demonstrate the role of the contrastive causal-chain explanation stage. The representative frames show when the maneuver prediction emerges over time, while the causal-chain diagrams explain how the observed states support the selected maneuver and suppress the alternatives. This directly connects the temporal prediction behavior to the learned causal structure and to the local intervention-based explanation mechanism. Therefore, the framework provides sample-level explanations of why RLC or LLC is favored, why LK is less supported, and why the opposite lane-change direction is not selected in the corresponding scenario.

\noindent\textbf{Supplementary videos.}
The full temporal evolution of the two qualitative examples is provided as supplementary videos. In addition, an accompanying playlist includes two further qualitative examples to show the prediction and explanation behavior in additional driving situations. The corresponding links are stated in \Cref{tab:supplementary-videos}.
\vspace{-2mm}
\begin{table}[pos=!htbp]
\centering
\caption{Supplementary video material for the qualitative causal-chain examples.}
\label{tab:supplementary-videos}
\setlength{\tabcolsep}{5pt}
\begin{tabular}{|
>{\centering\arraybackslash}m{0.18\textwidth}|
>{\centering\arraybackslash}m{0.36\textwidth}|
>{\centering\arraybackslash}m{0.38\textwidth}|}
\hline
\textbf{Scenario} & \textbf{Content} & \textbf{Link} \\
\hline
Right Lane Change
& Full temporal evolution of the RLC example shown in \Cref{fig:qualitative-rlc-frames}.
& \url{https://youtu.be/hqUIwALstuw} \\
\hline
Left Lane Change
& Full temporal evolution of the LLC example shown in \Cref{fig:qualitative-llc-frames}.
& \url{https://youtu.be/vCPD0ok8eWE} \\
\hline
Additional qualitative examples
& Playlist containing additional videos for further qualitative inspection.
& \url{https://www.youtube.com/playlist?list=PLAeK3AuwxenHU5x5tALkMscj-mbkKYVEu} \\
\hline
\end{tabular}
\vspace{-5mm}
\end{table}

\vspace{-5mm}
\section{Conclusion}
\label{sec:conclusion}

This work presented a causal-inference-based framework for lane-change prediction and explanation. The proposed approach converts numerical driving variables into interpretable linguistic states, incorporates expert knowledge through a constraint matrix, and learns an expert-constrained \ac{SCM} using DECI. This allows the framework to model dependencies among traffic context, vehicle motion, gap, relative velocity, TTC, and the future maneuver, instead of treating all inputs as independent predictors. The results show that the learned causal structure follows physically meaningful relations and that intervention analysis can identify variables with strong or weak effects on the maneuver probabilities. Lateral-motion and interaction-related variables produced the strongest intervention effects, while refutation tests showed that selected effects remained stable under placebo, random-common-cause, and data-subset diagnostics. The trained DECI-based \ac{SCM} also achieved strong prediction performance close to the lane-marking crossing event, with macro F1-scores above 95\% during the first three seconds before crossing, while still providing maneuver anticipation in the 7--8 s interval before the lane-change event. In addition to prediction accuracy, the framework provides contrastive causal-chain explanations for individual samples. The qualitative examples show how the model traces supporting paths for the predicted maneuver and identifies why alternative maneuvers are less supported. A diagnostic ablation experiment further showed that variables estimated to have weak intervention effects can be removed with limited impact on prediction performance, supporting the practical value of the intervention analysis. Overall, the proposed framework combines maneuver prediction, causal-effect analysis, robustness checking, and causal-chain explanation within a unified \ac{SCM}. Future work will extend the framework to additional datasets and driving environments, incorporate richer temporal causal structures, investigate alternative methods for linguistic feature categorization, and test the model on other safety-related tasks, such as decision-making support and risk-aware maneuver planning.


\bibliographystyle{cas-model2-names}

\bibliography{manzour-et-al-references}


\newpage

\bio{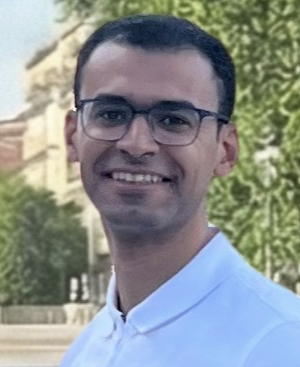}
Mohamed Manzour Hussien obtained his bachelor’s degree in Mechatronics from the German University in Cairo (GUC) in 2019. He later worked as a Lecturer Assistant at GUC and completed his master’s degree in Intelligent Transportation Systems (ITS) in 2022 at the Multi-Robot Systems (MRS) research group, focusing on pedestrian behavior prediction. In 2023, he started his Ph.D. in Intelligent Transportation Systems at the INtelligent VEhicles and Traffic Technologies (INVETT) research group, University of Alcalá, Spain. His doctoral research focused on vehicle behavior prediction, particularly lane change prediction. In the same year, he won first place in the IEEE ITSS Student Competition in the Driver Decision Prediction track. He completed his Ph.D. in 2026 and is currently a postdoctoral researcher at the University of Alcalá, continuing his research in intelligent vehicles.
\endbio

\vspace{5mm}

\bio{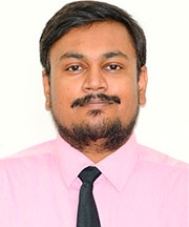}
Aditya Kumar is a PhD candidate with the Intelligent Transportation Systems at the INtelligent VEhicles and Traffic Technologies (INVETT) research group, University of Alcalá, Spain, and currently pursuing his research in the field of intelligent vehicles. In 2024, he graduated with a Master of Technology in Computer Sciences and Technology from Jawaharlal Nehru University (JNU). The IIIT Delhi CHANAKYA Fellowship funded his research during his Master's program in the area of detecting various road anomalies present in unstructured road environments. Aditya Kumar had previously obtained a degree from Guru Gobind Singh Indraprastha University (GGSIPU) in 2017 in Bachelor of Technology in Electronics and Communication. \vspace{30mm}
\endbio

\vspace{-23mm}
\bio{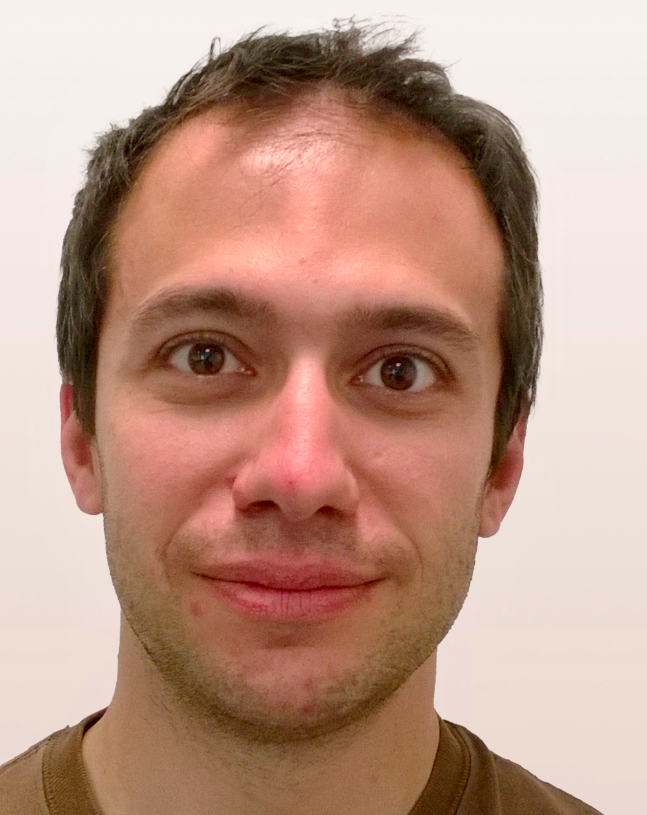}
Augusto Luis Ballardini was born in Buenos Aires, Argentina, in 1984. He completed his M.Sc. and Ph.D. degrees in Computer Science from the Università degli Studi di Milano - Bicocca, Italy, in 2012 and 2017 respectively. Following his post-doctoral activities in the IRALAB Research Group for two years, Dr. Ballardini joined the INVETT Research Group at the Universidad de Alcalá, Spain, in 2019. During his time at INVETT, he was awarded a Marie Skłodowska-Curie Actions research grant and a research grant within the Maria Zambrano/NextGenerationEU project from the Spanish Ministry of Science, Innovation, and Universities. His research focuses on developing advanced systems for autonomous vehicle localization and data fusion, using heterogeneous data sources such as digital maps, LiDAR, and image data, combined with cutting-edge computer vision and machine learning algorithms.
\endbio
\vspace{5mm}
\bio{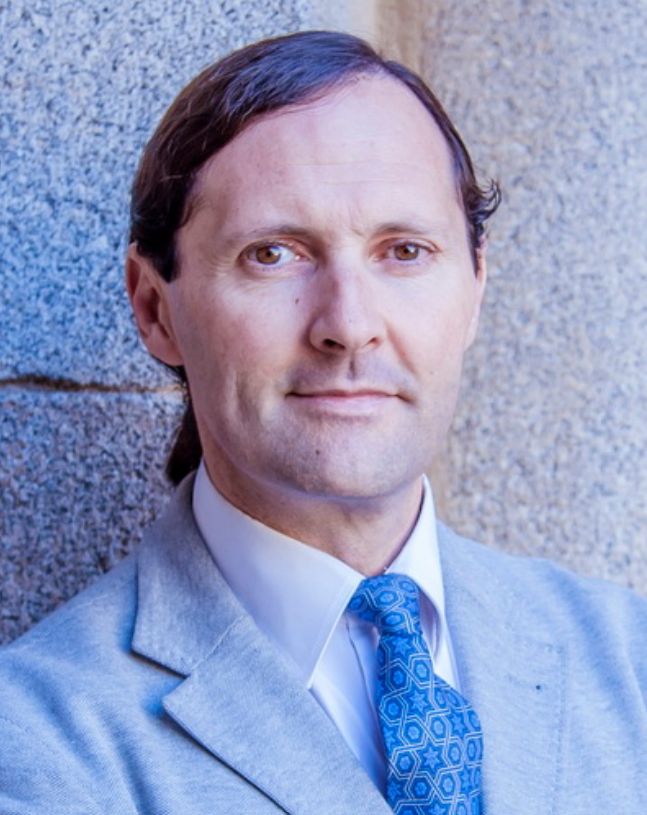}
Miguel \'Angel Sotelo received the degree in Electrical Engineering in 1996 from the Technical University of Madrid, the Ph.D. degree in Electrical Engineering in 2001 from the University of Alcalá (Alcalá de Henares, Madrid), Spain, and the Master in Business Administration (MBA) from the European Business School in 2008. He is currently a Full Professor at the Department of Computer Engineering of the University of Alcalá (UAH). His research interests include Self-driving cars, Prediction Systems, and Traffic Technologies. He is author of more than 300 publications in journals, conferences, and book chapters. He has been recipient of the Best Research Award in the domain of Automotive and Vehicle Applications in Spain in 2002 and 2009, and the 3M Foundation Awards in the category of eSafety in 2004 and 2009. Miguel Ángel Sotelo has served as Project Evaluator, Rapporteur, and Reviewer for the European Commission in the field of ICT for Intelligent Vehicles and Cooperative Systems in FP6 and FP7. He was Editor-in-Chief of the IEEE Intelligent Transportation Systems Magazine (2014-2016), Associate Editor of IEEE Transactions on Intelligent Transportation Systems (2008-2014), member of the Steering Committee of the IEEE Transactions on Intelligent Vehicles (since 2015), and a member of the Editorial Board of The Open Transportation Journal (2006-2015). He has served as General Chair of the 2012 IEEE Intelligent Vehicles Symposium (IV’2012) that was held in Alcalá de Henares (Spain) in June 2012. He was recipient of the IEEE ITS Outstanding Research Award in 2022, the IEEE ITS Outstanding Application Award in 2013, and the Prize to the Best Team with Full Automation in GCDC 2016. He is a Former President of the IEEE Intelligent Transportation Systems Society.
\endbio

\end{document}